\documentclass[conference,compsoc]{IEEEtran}
\IEEEoverridecommandlockouts
\usepackage[table]{xcolor}
\usepackage{cite}
\usepackage{amsmath,amssymb,amsfonts}
\usepackage{graphicx}
\usepackage{textcomp}

\usepackage{xspace}
\usepackage{lipsum}
\usepackage{subfigure} 

\usepackage{color, xcolor}
\usepackage[linesnumbered,ruled]{algorithm2e}
\usepackage{booktabs}
\usepackage{multirow}
\usepackage{threeparttable}
\usepackage{array}
\usepackage{bbding}

\usepackage{epigraph} 
\usepackage{balance}

\usepackage{enumitem}

\usepackage{breakurl}
\usepackage[breaklinks]{hyperref}

\hypersetup{
    colorlinks=true,
    linkcolor=[rgb]{0.2 0.74 0.0},
    urlcolor=[rgb]{0.18 0.5 0.9},
    citecolor=[rgb]{0.73 0 0.05},
}

\footnoterule
\renewcommand{\footnoterule}{%
  \kern -4.5pt
  \hrule width .2\textwidth height 0.5pt
  \kern 4pt
}

\newcommand{\sys}{{\sc Sophon}\xspace}

\begin{document}

\title{{\LARGE \sys}: Non-Fine-Tunable Learning to Restrain Task Transferability \\ For Pre-trained Models\\
}


\author{\IEEEauthorblockN{Jiangyi Deng\textsuperscript{1}\IEEEauthorrefmark{1}\thanks{\IEEEauthorrefmark{1}Equal Contribution}, Shengyuan Pang\textsuperscript{1}\IEEEauthorrefmark{1}, Yanjiao Chen\textsuperscript{1}, Liangming Xia\textsuperscript{1}, \\Yijie Bai\textsuperscript{1}, Haiqin Weng\textsuperscript{2}, Wenyuan Xu\textsuperscript{1}}
\IEEEauthorblockA{\textsuperscript{1}\textit{Zhejiang University}, \textsuperscript{2}\textit{Ant Group}
}
%
}

\maketitle

\begin{abstract}

Instead of building deep learning models from scratch, developers are more and more relying on adapting pre-trained models to their customized tasks. However, powerful pre-trained models may be misused for unethical or illegal tasks, \textit{e.g.}, privacy inference and unsafe content generation. 
In this paper, we introduce a pioneering learning paradigm, \textit{non-fine-tunable learning}, which prevents the pre-trained model from being fine-tuned to indecent tasks while preserving its 
performance on the original task.
To fulfill this goal, we propose \sys, a protection framework that reinforces a given pre-trained model to be resistant to being fine-tuned in pre-defined restricted domains. Nonetheless, this is challenging due to a diversity of complicated fine-tuning strategies that may be adopted by adversaries. Inspired by model-agnostic meta-learning, we overcome this difficulty by designing sophisticated fine-tuning simulation and fine-tuning evaluation algorithms. In addition, we carefully design the optimization process to entrap the pre-trained model within a hard-to-escape local optimum regarding restricted domains. We have conducted extensive experiments on two deep learning modes (classification and generation), seven restricted domains, and six model architectures to verify the effectiveness of \sys. Experiment results verify that fine-tuning \sys-protected models incurs an overhead comparable to or even greater than training from scratch. Furthermore, we confirm the robustness of \sys to three fine-tuning methods, five optimizers, various learning rates and batch sizes. \sys may help boost further investigations into safe and responsible AI.

\end{abstract}

\section{Introduction}

\begin{figure}[t]
    \centering
\setlength{\abovecaptionskip}{0pt}
\setlength{\belowcaptionskip}{0pt}


\includegraphics[width=3.4in, trim=280 90 280 90, clip]{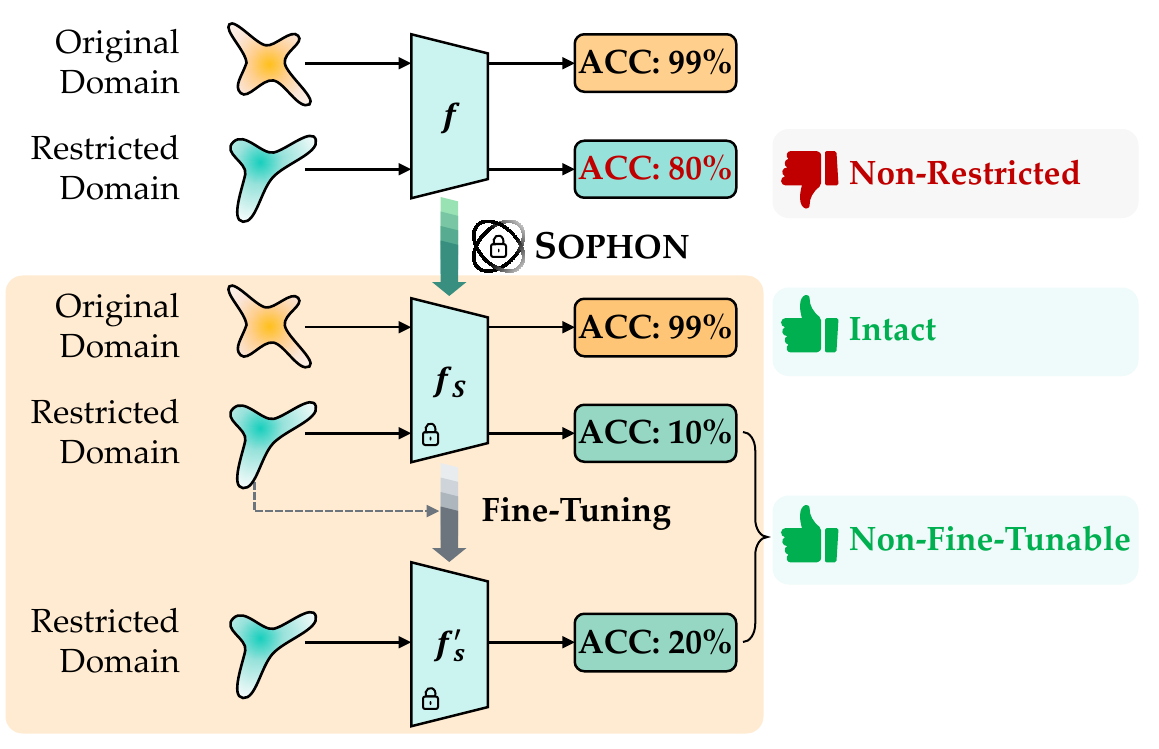}
\raggedbottom

\caption{The objectives of non-fine-tunable learning. (1) Intactness: it should preserve the model performance in the original domain. (2) Non-fine-tunability: fine-tuning the model in the restricted domain should incur a comparable or even greater overhead than training the model from scratch.}\label{fig:goal}
\end{figure}

Pre-trained models
have become a popularity for developers to handle the increasingly complex
and diverse deep learning tasks. A pre-trained model learns universal representations from large-scale datasets with high generalization ability. By simply fine-tuning it with
a few task-specific data samples, a user can transfer the pre-trained model to different downstream
tasks~\cite{HeZRS16,SermanetEZMFL13,RenHGS15,LongSD15,VinyalsTBE15}, which is significantly more efficient
than training models from scratch. Attracted by this, a plethora of pre-trained models have been built and available in public
model zoos to facilitate the research community and industry. 

However, this emerging pre-training solution may be misused to unethical or harmful tasks~\cite{bengio2023managing}, \textit{e.g.}, using classification models to predict sensitive information like sexual orientation~\cite{Unethical, Pregnant, employee, facebook} or generative models to create unsafe content like sexually-explicit, violent, and political images~\cite{disturbing, AI_created_child, AI_porn, AI_porn_easy, Paedophiles}. Although efforts are being made to train safe and  ethical models from scratch~\cite{stop_unethical_ai, factsheet}, high transferability of pre-trained models makes them more vulnerable to being abused. Non-transferable learning (NTL)~\cite{wang2022non, wang2023model, zeng2022unsupervised, wang2023domain} has been proposed to degrade the performance of a pre-trained model in certain domains. However, non-transferable learning only considers the transferability of a pre-trained model \emph{before fine-tuning}. After being fine-tuned with only a few samples from the target domain, an NTL-protected model will be easily transformed into a well-performed model for the target task. As far as we are concerned, there is a lack of effective protection for pre-trained models from being fine-tuned to inappropriate tasks. 

To bridge this gap, we propose a new learning paradigm, dubbed \emph{non-fine-tunable learning}, which aims to realize two objectives, as illustrated in Figure~\ref{fig:goal}. (1)~\textbf{Intactness}. 
Non-fine-tunable learning should preserve the model performance in the original domain.
 (2)~\textbf{Non-Fine-Tunability}. Fine-tuning the pre-trained model in restricted domains should incur a comparable or even greater overhead than training the model for restricted tasks from scratch. Note that non-transferable learning is a special case of non-fine-tunable learning with an identity function as the fine-tuning process.

 However, to materialize these goals are challenging in several aspects.


\begin{itemize}
\setlength{\itemsep}{0pt}
    \item \textit{How to design the optimization framework to obtain the two objectives of non-fine-tunable learning?}
\end{itemize}

The two objectives of intactness and non-fine-tunability should be instantiated in the optimization framework. To ensure non-fine-tunability, we need to estimate the model performance under hypothetical fine-tuning processes. However, commonly-used fine-tuning methods are iterative and complicated optimization problems themselves, making it difficult to integrate them into the optimization framework. To address this challenge, we are inspired by the model-agnostic meta-learning (MAML) framework~\cite{finn2017model} to simulate the fine-tuning process of a potential adversary. In this way, we formulate the non-fine-tunable learning problem into a multi-objective optimization framework, with a fine-tuning suppression term and a normal training reinforcement term. The former minimizes the model performance in restricted domains with the feedback provided by the simulated fine-tuning processes. The latter maximizes the model performance in the original domain to ensure intactness. 

\begin{itemize}
\setlength{\itemsep}{0pt}
    \item \textit{How to ensure robustness under unpredictable fine-tuning strategies adopted by adversaries?}
\end{itemize}

A non-cooperative adversary may fine-tune the pre-trained model with unpredictable fine-tuning strategies. There are tons of choices for fine-tuning strategies, including model initialization, transfer strategy and optimization strategy. The simulated fine-tuning process in our optimization framework may not coincide with that adopted by the adversary. To tackle this problem, we integrate various empirically-strong fine-tuning strategies in the fine-tuning simulation process. 
Our extensive experiments have confirmed that our simulated fine-tuning strategies are effective in ensuring non-fine-tunability to unseen fine-tuning strategies.

\begin{itemize}
\setlength{\itemsep}{0pt}
    \item \textit{How to boost the convergence of fine-tuning suppression in restricted domains?}
\end{itemize}

Gradient descent is commonly used to solve the optimization problem in deep learning. To guarantee convergence of the gradient descent algorithm, the gradient value used to update model parameters should decrease with the training process. Conventional loss functions, \emph{e.g.}, cross-entropy loss are designed to satisfy this condition. However, to minimize the model performance in restricted
domains is opposite to normal training that tries to maximize the model performance on training datasets. Therefore, conventional loss functions destabilize the convergence of solving our optimization problem. To resolve this difficulty, we propose two alternative loss functions for classification and one for generation, \emph{i.e.}, inverse cross-entropy, Kullback–Leibler divergence from uniform distribution and denial of service losses, gradients of which are theoretically shown to decrease with the iterative process of fine-tuning suppression. The design of these loss functions facilitates the convergence of fine-tuning suppression in restricted domains.

We implement a fully-functional non-fine-tunable prototype named \sys\footnote{Sophon is the name of a supercomputer created by Trisolarans (aliens) to sabotage Earth's technological development in the science fiction \textit{The Three-Body Problem}~\cite{liu2014three} by Cixin Liu.}, which conducts non-fine-tunable learning on a given pre-trained model before releasing it to prevent its use or adaptation for restricted tasks. Extensive experiments have been conducted to evaluate the performance of \sys on two deep learning modes (classification and generation), seven restricted tasks, and six model architectures. We demonstrate that fine-tuning \sys-protected models has overhead close to or even greater than training from scratch, under three fine-tuning methods, five optimizers, various learning rates and batch sizes. We have open-sourced our code\footnote{\url{https://github.com/ChiangE/Sophon}} in a hope to incentivize more research in this area.

\vspace{5pt}
\noindent\textbf{Summarization of contributions.} 
We summarize our theoretical and technical contributions as follows:
\begin{itemize}
\setlength{\itemsep}{5pt}
    \item We propose non-fine-tunable learning, which aims to restrain transferability of pre-trained models to improper tasks under fine-tuning, extending the connotation of safe and ethic deep learning models.
    
    \item We develop a non-fine-tunable learning framework, which optimizes pre-trained models to fulfill the objectives of intactness and non-fine-tunability, which may help reduce the risk of model abuse for the open source community.
    
    \item We conduct extensive experiments to verify the effectiveness and robustness of our non-fine-tunable learning framework against a diversity of fine-tuning strategies.
\end{itemize}

\section{Background}

\subsection{Deep Learning Tasks}\label{subsec:deeplearning}

In this paper, we focus on two major tasks of deep learning, \textit{i.e.}, classification and generation. They have different paradigms including model architectures, loss functions, and datasets.

\emph{Classification} tasks aim to predict the class label of an input sample, \textit{e.g.}, classifying emails as spam or not spam.  A deep learning classification model $f\left(x\right)=\sigma\left(z_{\theta}\left(x\right)\right)$ usually consists of a model backbone $z_{\theta}\left(\cdot\right)$ and a \texttt{Softmax} layer $\sigma\left(\cdot\right)$, where $z_{\theta}$ is parameterized by $\theta$.
Typically, the output $y$ is a probability distribution as $y=\left(y_{1}, y_{2}, \cdots, y_{C}\right)$, where $C$ is the number of classes and the predicted label is $\arg\max_k y_{k}$.  Given a training dataset
$\mathcal{D}=\left\{\left(x_i,y_i\right)|x_i\in\mathcal{X}, y_i\in\mathcal{Y}, i=1,2,\cdots\right\}$, a classification model learns a function $f:\mathcal{X}\rightarrow \mathcal{Y}$ by minimizing the loss function $\mathcal{L}\left(f(x_i),y_i\right)$, where $y_i$ is the ground-truth probability distribution.
A commonly used loss function is the cross-entropy loss $\mathcal{L}_{\mathrm{CE}}=-\frac{1}{|\mathcal{\mathcal{X}}|}\sum_{x_i\in \mathcal{X}} y_{i}\log f(x_i)$.


\emph{Generation} tasks aim to generate data samples that follow a certain distribution. Representative generative models include autoencoder (AE)~\cite{rumelhart1986learning}, variational autoencoder (VAE)~\cite{KingmaW13}, generative adversarial network (GAN)~\cite{GoodfellowPMXWOCB14}, and diffusion probabilistic model~\cite{ho20denoising, rombach22high}. Without loss of generality, we focus on diffusion probabilistic models, the state-of-the-art generative models. But note that \sys can be easily extended to other generative models. Diffusion probabilistic models (referred to as diffusion models for brevity) is a parameterized Markov chain that generally contains a \emph{forward process} and a \emph{reverse process}~\cite{ho20denoising, rombach22high}. The \emph{forward process} or \emph{diffusion process} adds Gaussian noises to data and the \emph{reverse process} are learned to reverse the diffusion process, \textit{i.e.}, denoise the data. To train a diffusion model, training datasets are crafted as $\mathcal{D}=\left\{\left((x_t,t), \epsilon\right)|x_t\sim\mathcal{X}_t, t \in [1, T], \epsilon\sim\mathcal{N}(\mathbf{0}, \mathbf{I})\right\}$, where $x_t\left(x_0, \epsilon, t\right) = \sqrt{\bar{\alpha}_t}\,x_0 + \sqrt{1-\bar{\alpha}_t}\,\epsilon$ is the forward process with predefined hyper-parameters $\bar{\alpha}_t$, $\mathcal{X}_t$ is the distribution of $x_t$, and $T$ is the number of diffusion steps. 
The diffusion model $f:\mathcal{X}_t\rightarrow \mathcal{N}, \forall t \in [1,T]$ learns to predict noises at each diffusion step by minimizing the loss function $\mathcal{L}\left(f(x_{ti}, t_i\right), \epsilon_i)$. A commonly used loss function is the mean squared error  $\mathcal{L}_{\mathrm{MSE}}=\frac{1}{|\mathcal{X}|}\sum_i\|\epsilon_i-f(x_{ti}, t_i)\|^2$.


\subsection{Transfer Learning}


Transfer learning is a learning paradigm that enables knowledge transfer across various learning tasks. The intuition behind transfer learning is to encode knowledge from a large and general enough dataset in a pre-trained model, and then fine-tune the pre-trained model on specific tasks~\cite{Bengio12Deep}. Following this intuition, transfer learning formalizes a two-phase learning framework: a pre-training phase to acquire knowledge from one or more source tasks, and a fine-tuning stage to transfer the attained knowledge to target tasks. Owing to the wealth of knowledge captured in the pre-training phase, the fine-tuning phase can quickly adapt models to new tasks with limited training samples and training efforts~\cite{yu2022metaformer, Devlin19Bert, ren15faster}. 

As deep learning models grow larger and more complicated, fine-tuning large-scale pre-trained models instead of learning models from scratch has become a consensus. A large quantity of pre-trained models are developed by individuals or tech companies like Google, Meta, Microsoft and NVIDIA, and are released as open-sourced models on model hubs (\textit{e.g.}, GitHub and Hugging Face), allowing developers to fine-tune these models into their customized applications.

The fine-tuning process adapts pre-trained models to downstream tasks. The key components of fine-tuning include model initialization, transfer strategy and optimization strategy.

\textit{\underline{Model initialization}}. Given a pre-trained model, a customized model may be initialized in various ways, \textit{e.g.}, directly using the entire pre-trained model, copying the entire model but randomly initializing the last fully-connected layer, and copying the entire model but randomly initializing the last few layers. By copying more layers from the pre-trained model, more knowledge will be inherited. 

\textit{\underline{Transfer strategy}}.
The transfer strategy selects which parameters to update during fine-tuning. Two common strategies are fine-tuning the whole model or only fine-tuning certain layers. Previous study shows that fine-tuning the whole model yields better performance than only fine-tuning a few layers, due to a gain of co-adaptation within layers~\cite{YosinskiCBL14}.
    
\textit{\underline{Optimization strategy}}. The optimization strategy includes the optimizer (the parameter update rules) and hyper-parameters (\textit{e.g.}, batch size, learning rate, and the number of updates). Commonly used optimizers in deep learning include SGD~\cite{robbins1951stochastic}, Momentum~\cite{polyak1964some}, Nesterov~\cite{SutskeverMDH13}, Adagrad~\cite{DuchiHS11}, Adadelta~\cite{Zeiler2012adadelta}, and Adam~\cite{kingma2014adam}.

To prevent pre-training models from being abused for inappropriate downstream tasks, non-transferable learning (NTL)~\cite{wang2022non, wang2023model, zeng2022unsupervised, wang2023domain} was proposed to degrade the performance of pre-trained models in certain tasks (referred to as restricted domain) before fine-tuning. As far as we are concerned, existing works on non-transferable learning all focus on classification tasks. Non-transferable learning incorporates additional optimization objectives, \textit{e.g.}, mis-classifying samples from the restricted domain and enlarging distance between representation distributions of the source and the restricted domains. However, non-transferable learning does not suppress the performance of pre-trained models in the restricted domain if fine-tuning is performed. In other words, fine-tuning NTL-based pre-trained model will break performance suppression and attain high-quality performance in the restricted domain. We will demonstrate that NTL cannot resist fine-tuning via experiments in \S\ref{subsec:effectiveness}.



\begin{figure*}[t]
    \centering
\setlength{\abovecaptionskip}{2pt}
\setlength{\belowcaptionskip}{0pt}

\includegraphics[width=7.in, trim=30 25 30 20, clip]{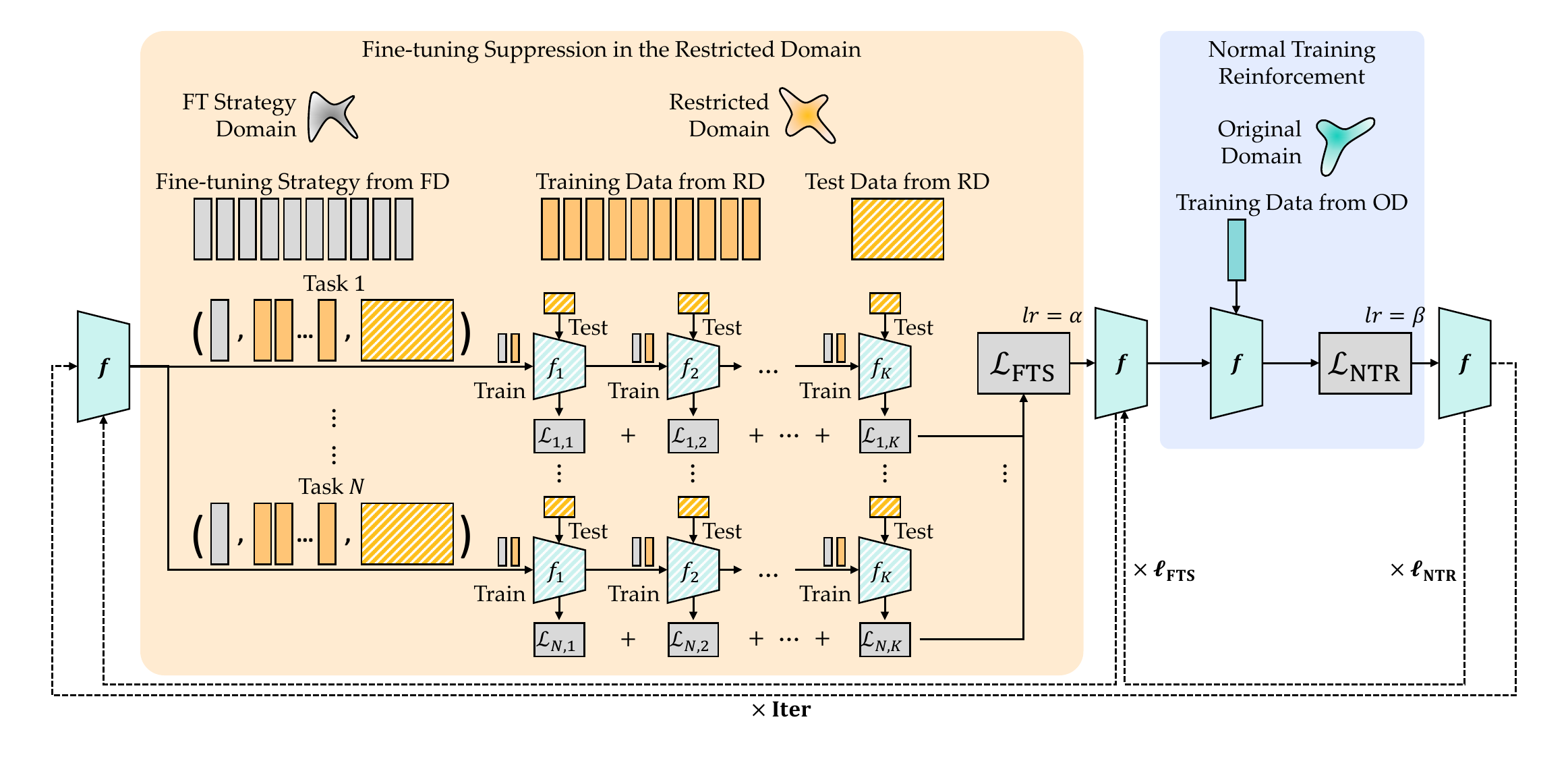}

\caption{Design of \sys. \sys mainly consists of two alternating phases, \textit{i.e.}, the fine-tuning suppression (FTS) loops and the normal training reinforcement (NTR) loops. The FTS loops are designed to simulate different fine-tuning processes and degrade the fine-tuning performance in the restricted domain. The NTR loops are designed to maintain the performance in the original domain. The number of tasks $N$, the number of updates $K$, the learning rates of FTS loops $\alpha$ and NTR loops $\beta$, and the number of FTS loops $\ell_\mathrm{FTS}$ and NTR loops $\ell_\mathrm{NTR}$, and the total number of iterations $\mathrm{Iter}$ are hyper-parameters.}\label{fig:workflow}
\end{figure*}

\subsection{Design Goal}\label{subsec:goal}
We first define the system model in terms of the adversary and defender, and then elaborate the design goals of \sys under the defined system model.

\emph{Adversary.}  The adversary aims to obtain a well-performed model in the restricted domain, \textit{e.g.}, sensitive information inference or unsafe image generation. To achieve this objective, the adversary fine-tunes an open-sourced public pre-trained model on data samples from the restricted domain. The adversary may be informed that the model is \sys-protected and can leverage any fine-tuning strategies to try to neutralize \sys.

\emph{Defender.} The defender aims to prevent a pre-trained model from being fine-tuned for certain downstream tasks. The defender defines the restricted domain according to well-recognized unethical applications and can glean samples from the restricted domain. The defender has access to the pre-trained model and can modify the pre-trained model.

Given the system model, we lay out three major goals of \sys.

    \textbf{Intactness}. The protected model should perform as well as the original model on benign tasks. Let $\mathcal{D}_S$ denotes the source-domain datasets used to train the pre-trained model. Intactness indicates that $\mathbb{E}_{x\sim \mathcal{D}_S}\ \mathcal{L}(f_{\theta}(x))\approx \mathbb{E}_{x\sim \mathcal{D}_S}\ \mathcal{L}(f_{0}(x))$, where $f_{0}$ and $f_{\theta}$ denote the original model and the protected model respectively.
    
    \textbf{Non-transferability}. The protected model (without fine-tuning) should perform poorly in the restricted domain. Let $\mathcal{D}_A$ denotes the datasets from the restricted domain. Non-transferability indicates $\mathbb{E}_{x\sim \mathcal{D}_A}\ \mathcal{L}(f_\theta(x))\gg \mathbb{E}_{x\sim \mathcal{D}_A}\ \mathcal{L}(\phi(f_0(x)))$, where $\phi(\cdot)$ is the fine-tuning strategy.
    
    \textbf{Non-fine-tunability}. The protected model, after being fine-tuned by the adversary, should still perform poorly in the restricted domain, \textit{i.e.}, $\mathbb{E}_{x\sim \mathcal{D}_A}\ \mathcal{L}(\phi(f_\theta(x))\gg \mathbb{E}_{x\sim \mathcal{D}_A}\ \mathcal{L}(\phi(f_0(x)))$. Note that non-transferability is a special case of non-fine-tunability, since the left-hand-side $\phi$ can be an identity function.

Note that, if fine-tuning the protected model results in poorer performance than training from scratch, the adversary can simply train a model from scratch using data from the restricted domain. Thus, the worst-case scenario for the adversary is training from scratch. If fine-tuning the protected model achieves better performance than training from scratch but poorer than fine-tuning the original model, we consider the adversary to have achieved partial success.


\section{Problem Formulation of Non-Fine-Tunable Learning}\label{sec:nftl}

Before delving into the design details of \sys, in this
section, we formally formulate the non-fine-tunable learning as a
constrained optimization problem.

Non-fine-tunable learning aims to optimize a model parameterized by $\theta$ that satisfies the design goals of intactness, non-transferability and non-fine-tunability.

\vspace{5pt}
\noindent\underline{\textbf{Basic Formulation:}}
\begin{equation}\label{equ:obj1}
\begin{aligned}
\mathop{\min}\limits_{\theta}\ &-\mathbb{E}_{x\sim \mathcal{D}_A,\phi\sim\Phi}\ \mathcal{L}\left(\phi\left(f_\theta\left(x\right)\right)\right),\\
\ \ \mathrm{s.t.}~&~\mathbb{E}_{x\sim \mathcal{D}_S}\left(\max\left\{0, \mathcal{L}\left(f_\theta\left(x\right)\right)-\mathcal{L}\left(f_0\left(x\right)\right)\right\}\right)<\lambda,
\end{aligned}
\end{equation}
\noindent where $\lambda$ denotes the tolerance of the performance degradation in the original domain. The constrained optimization problem in Equation~\eqref{equ:obj1} can be very difficult to solve. Thus, we instead solve the following unconstrained optimization problem.

\vspace{5pt}
\noindent\underline{\textbf{\sys Formulation:}}
\begin{equation}\label{equ:obj2}
\begin{aligned}
\mathop{\min}\limits_{\theta}\ &-\mathbb{E}_{x\sim\mathcal{D}_A,\phi\sim\Phi}\ \mathcal{L}\left(\phi\left(f_\theta\left(x\right)\right)\right)+\mu\cdot\mathbb{E}_{x\sim\mathcal{D}_S}\ \mathcal{L}\left(f_\theta\left(x\right)\right),
\end{aligned}
\end{equation}
\noindent where $\mu$ balances the two design goals of intactness and non-fine-tunability. Nonetheless, the problem in Equation~\eqref{equ:obj2} is still hard to solve since the fine-tuning process $\phi$ is usually iterative and does not have a closed-form solution. Thus, we propose \sys to derive an approximate solution of Equation~\eqref{equ:obj2}.

\section{\sys: Design Details}\label{sec:design}

As shown in Figure~\ref{fig:workflow}, \sys consists of two key optimization modules, \textit{i.e.}, fine-tuning suppression in the restricted domain and normal training reinforcement in the original domain. The fine-tuning suppression module is designed to degrade the fine-tuning performance in the restricted domain in simulated fine-tuning processes. The normal training reinforcement module is designed to maintain the performance in the original domain. 


\subsection{Fine-Tuning Suppression in Restricted Domain}
Fine-tuning suppression in the restricted domain corresponds to the first term in Equation~\eqref{equ:obj2}, which is hard to optimize for two reasons. First, the fine-tuning process $\phi(f(x))$ is an optimization problem itself, and does not have a closed-form solution. Thus, $\mathcal{L}(\phi(f_\theta(x)))$ cannot be computed directly. Second, to the best of our knowledge, there is no existing closed-form metric that reliably implies the performance of a model after fine-tuning, \textit{i.e.}, a metric that positively correlates to $\mathcal{L}(\phi(f_\theta(x)))$. Thus, there is no feedback signal for optimization.
Inspired by the model-agnostic meta-learning (MAML) framework~\cite{finn2017model}, we design simulated fine-tuning processes to approximate the model performance after fine-tuning, which serves as the feedback for optimization.

\subsubsection{Fine-Tuning Simulation}

We denote a fine-tuning environment as a triplet $\left(\phi_i, \mathcal{R}_i, \mathcal{T}_i \right)$, where $\phi_i(\cdot)\sim\Phi$ is a fine-tuning strategy, $\mathcal{R}_i$ is the dataset used to fine-tune model $f_\theta$, and $\mathcal{T}_i$ is the dataset used to evaluate the performance of model $f_\vartheta^k$ after the $k$-th round of fine-tuning. Both $\mathcal{R}_i$ and $\mathcal{T}_i$ are sampled from the restricted domain. 

At the $k$-th round, the fine-tuning process is essentially an optimization problem. 
\begin{equation}\label{equ:phi}
\mathop{\max}\limits_{\vartheta}\ \mathbb{E}_{x\sim\mathcal{R}_i}\ \mathcal{L}\left(f_\vartheta^k\left(x|\phi_i, f_\vartheta^{k-1}\right)\right),
\end{equation}
where $f_\vartheta^k$ is the fine-tuned version of $f_\vartheta^{k-1}$. Note that $f_\vartheta^0 = f_\theta$.

The fine-tuning process may not converge, \textit{i.e.}, $K\rightarrow \infty$, where $K$ is the total number of fine-tuning rounds. Thus, we set an upper-bound on $K$ as a relatively large integer to approximate the point of convergence, \textit{e.g.}, 50 in our experiments. 


We consider the strongest adversary with the (potentially) best fine-tuning strategy to try to obtain a good performance in the restricted domain. Therefore, we materialize $\phi_i$ as follows.  For model initialization, we choose two strategies. The first one uses the entire pre-trained model, and the second one uses the entire pre-trained model except for the final layer. For transfer strategy, we choose to update the whole model, and for the optimization strategy, we use the Adam optimizer.

\subsubsection{Fine-tuning Evaluation} 

The fine-tuning is simulated on behalf of the adversary and the defender aims to suppress the fine-tuning performance. Therefore, we propose to use the aggregate error of all fine-tuned models generated at every iteration by every fine-tuning strategy on the test dataset to evaluate potential fine-tuning performance of a pre-trained model $f_\theta$. The first term of Equation~\eqref{equ:obj2}, defined as $\mathcal{L}_{\mathrm{FTS}}$, is expressed as
\begin{equation}\label{equ:nftloss}
\begin{aligned}
\mathcal{L}_\mathrm{FTS} =  \sum_{i=1}^N\sum_{k=1}^K\, \gamma_{i,k}\cdot\mathcal{L}_\alpha\left(f_{\vartheta}^k|\phi_i, \mathcal{T}_{i}\right),
\end{aligned}
\end{equation}
where $\mathcal{L}_\alpha\left(f_{\vartheta}^k|\phi_i, \mathcal{T}_{i}\right)$ evaluate the performance of the protected pre-trained model fine-tuned by $\phi_i$ after $k$ iterations on the test dataset $\mathcal{T}_i$, and $\gamma_{i,k}$ is the weight. 

The protected pre-trained model is updated according to $\mathcal{L}_\mathrm{FTS}$ as
\begin{equation}\label{equ:nftupdate}
\begin{aligned}
\theta\leftarrow\theta\pm\alpha\cdot\nabla_{\theta}\mathcal{L}_\mathrm{FTS},
\end{aligned}
\end{equation}
where $\alpha$ denotes the learning rate. The sign of Equation~\eqref{equ:nftupdate} is determined by the design of the loss function $\mathcal{L}_\alpha$, which we elaborate in \S\ref{subsec:species}. Note that Equation~\eqref{equ:nftupdate} involves second-order gradient terms since $f_\vartheta^k$ is also optimized by gradient descent. To reduce computational complexity, we use the first-order approximation of Equation~\eqref{equ:nftupdate}~\cite{finn2017model} and update $\theta$ using an Adam optimizer~\cite{kingma2014adam}. 

\begin{algorithm}[tt]
\scriptsize
\caption{\sys}\label{alg:sophon}
\LinesNumbered
\KwIn{The original domain $\mathcal{S}$, the restricted domain $\mathcal{A}$, the fine-tuning setting domain $\Phi$.}
\KwIn{The original model parameter $\theta_0$, FTS learning rate $\alpha$, NTR learning rate $\beta$, the number of FTS loops $\ell_\mathrm{FTS}$, the number of NTR loops $\ell_\mathrm{NTR}$, the number of iterations $\mathrm{Iter}$.}

\KwOut{The non-fine-tunable model $\theta$}
\textbf{Initialize $\theta \gets \theta_0$.}

\For{$1$ \KwTo $\mathrm{Iter}$}{
    
    \textit{\# Fine-tuning suppression in the restricted domain}
    
    \For{$1$ \KwTo $\ell_\mathrm{FTS}$}{
    
    \For{$i\leftarrow 1$ \KwTo $N$}{
    \textbf{Sample} one fine-tuning setting $\phi_i\sim\Phi$
    
    \textbf{Sample} $K$ batches of $\mathcal{R}_i, \mathcal{T}_i\sim\mathcal{A}$.

    \For{$k\leftarrow 1$ \KwTo $K$}{
    
    \textbf{Fine-tune} $f_\vartheta^k\gets f_\vartheta^k\left(x|\phi_i, f_\vartheta^{k-1}\right)$
    
     \textbf{Compute}  $\mathcal{L}_{i,k}\leftarrow\mathcal{L}_\alpha\left(f_\vartheta^k|\phi_i, \mathcal{T}_i\right)$.
    }
    }
    $\mathcal{L}_\mathrm{FTS} \gets \sum_{i=1}^N\sum_{k=1}^K\, \gamma_{i,k}\cdot\mathcal{L}_{i,k}$
    
    \textbf{Update} $\theta \gets$ \texttt{Adam}($\theta$, $\nabla_\theta \mathcal{L}_\mathrm{FTS}$, $\alpha$)

    }
    
    \textit{\# Normal training reinforcement in the original domain}
    
    \For{$1$ \KwTo $\ell_\mathrm{NTR}$}{
    
    \textbf{Sample} a batch of $\mathcal{O}\sim\mathcal{S}$.

    \textbf{Compute} $\mathcal{L}_\mathrm{NTR} \gets \mathcal{L}_\beta\left(f_{\theta}| \mathcal{O}\right),$

    \textbf{Update} $\theta \gets$ \texttt{Adam}($\theta$, $\nabla_\theta \mathcal{L}_\mathrm{NTR}$, $\beta$)

    }
    
}

\end{algorithm}
\raggedbottom

\subsection{Normal Training Reinforcement in Original Domain}
The fine-tuning suppression module may affect the performance of the pre-trained model in the original domain. Therefore, we carry out normal training reinforcement to maintain the performance in the original domain, corresponding to the second term of Equation~\eqref{equ:obj2}, which is expressed as 
\begin{equation}\label{equ:ntloss}
\begin{aligned}
\mathcal{L}_\mathrm{NTR} = \mathcal{L}_\beta\left(f_{\theta}| \mathcal{O}\right),
\end{aligned}
\end{equation}
where $\mathcal{L}_\beta$ is the loss function used to measure the performance of protected pre-trained model in the original domain, and $\mathcal{O}$ is a training dataset sampled from the original domain. 

The protected pre-trained model is updated according to $\mathcal{L}_\mathrm{NTR}$ as
\begin{equation}\label{equ:ntupdate}
\begin{aligned}
\theta\leftarrow\theta-\beta\cdot\nabla_{\theta}\mathcal{L}_\mathrm{NTR},
\end{aligned}
\end{equation}
where $\beta$ denotes the learning rate. Equation~\eqref{equ:ntupdate} is also updated using an Adam optimizer~\cite{kingma2014adam}. We summarize the detailed process of solving Equation~\eqref{equ:obj2} in Algorithm~\ref{alg:sophon}.

\subsection{Instantiation of \sys}\label{subsec:species}

We instantiate model $f$ in Equation~\eqref{equ:obj2} as a classification model and a generative model as follows.
\subsubsection{Classification}\label{subsec:species:cls}

For classification tasks, the loss function $\mathcal{L}_\beta(\cdot)$ for normal training reinforcement can take the form of the traditional cross-entropy loss. However, using cross-entropy loss for the loss function $\mathcal{L}_\alpha(\cdot)$ of fine-tuning suppression is problematic due to the following reason. For the gradient descent process in Equation~\eqref{equ:nftupdate} to converge stably, the gradient term $\nabla\mathcal{L}_\mathrm{FTS}$ should decrease with the update iterations. Therefore, according to Equation~\eqref{equ:nftloss}, $\nabla\mathcal{L}_\alpha$ should decrease with the update iterations. Given our objective to amplify the error in the restricted domain with the update iterations, it is expected that the gradient $\nabla\mathcal{L}_\alpha$ should decrease as the error increases. This behavior is inherently linked to the nature of loss functions.
We theoretically analyze the nature of the cross-entropy loss, inspired by which we design two alternative loss functions.

As mentioned in \S\ref{subsec:deeplearning}, the output of a classifier can be represented as,
\begin{equation}\label{equ:cls}
\begin{aligned}
 &\left(\hat{y}_1, \hat{y}_2, \cdots, \hat{y}_C\right) = \sigma\left(z_1, z_2, \cdots, z_C\right),
\end{aligned}
\end{equation}
where $z_i$ is the $i$-th logit, $\hat{y}_i = \exp(z_i)/\sum_{j=1}^C\exp(z_j)$ is the predicted probability of the $i$-th label, $C$ is the total number of classes. Then, we have the following partial derivatives,
\begin{equation}\label{equ:partial}
\frac{\partial \hat{y}_j}{\partial z_i} = \left\{\begin{array}{rl}
       \hat{y}_i(1-\hat{y}_i), & i=j,\\
      -\hat{y}_i\hat{y}_j, & i\ne j.
\end{array}\right.
\end{equation}

Then, we can derive the gradient of the cross-entropy loss with respect to $z_i$ by the chain rule,
\begin{equation}\label{equ:CEgradient}
\begin{aligned}
\frac{\partial \mathcal{L}_{\mathrm{CE}}}{\partial z_i} =-\sum_{j=1}^{C}\frac{\partial \left(y_j\log \hat{y}_j\right)}{\partial \hat{y}_j}\cdot \frac{\partial \hat{y}_j}{\partial z_i} = \hat{y}_i - y_i.
\end{aligned}
\end{equation}

The magnitude of the gradient in Equation~\eqref{equ:CEgradient} is positively correlated with the error. In this case, the gradient in Equation~\eqref{equ:nftupdate} will increase with update iterations, thus the optimization process will easily diverge.

To address this problem, we propose two alternatives for the loss function $\mathcal{L}_\alpha$.

\textbf{Inverse Cross-Entropy (ICE).} We modify the cross-entropy loss into,
\begin{equation}\label{equ:ICE}
\begin{aligned}
\mathcal{L}_{\mathrm{ICE}}(f)=-\frac{1}{|\mathcal{X}|}\sum_i\sum_{j=1}^{C} y_{ij}\log\left(1-\hat{y}_{ij}\right).
\end{aligned}
\end{equation}
where $\hat{y}_{ij}$ is the $j$-th term of the prediction vector $y_i = f(x_i)$. 
Since increasing $-\log\left(\hat{y}_{ij}\right)$ is equivalent to decreasing $-\log\left(1-\hat{y}_{ij}\right)$, we can use the ICE loss to degrade the model performance in the restricted domain. Without loss of generality, we assume $y_1=1$. Then the gradient of the ICE loss with respect to $z_i$ is,
\begin{equation}\label{equ:ICEgradient}
\begin{aligned}
\frac{\partial \mathcal{L}_{\mathrm{ICE}}}{\partial z_i} &
= \left\{\begin{array}{rl}
       \hat{y}_1, & i=1,\\
      -\hat{y}_i\hat{y}_1/\left(1-\hat{y}_1\right), & i\ne 1.
\end{array}\right.
\end{aligned}
\end{equation}

From Equation~\eqref{equ:ICEgradient}, we know that the magnitude of gradient decreases as $\hat{y}_1$ decreases, \textit{i.e.}, $\lim_{\hat{y}_1\rightarrow 0} \left|\frac{\partial \mathcal{L}_{\mathrm{ICE}}}{\partial z_i}\right|=0.$
Therefore, the proposed inverse cross-entropy loss can better boost the convergence of fine-tuning suppression process.

\textbf{KL Divergence from Uniform Distribution (KLU).} The Kullback–Leibler (KL) divergence \cite{Kullback1951kl} measures the difference between two probability distributions. We leverage the KL divergence to force the output distribution of the model to be a uniform distribution, \textit{i.e.},
\begin{equation}\label{equ:KLU}
\begin{aligned}
\mathcal{L}_{\mathrm{KLU}}(f)=-\frac{1}{|\mathcal{X}|}\sum_i\sum_{j=1}^{C} \frac{1}{C}\log\left(C\cdot\hat{y}_{ij}\right).
\end{aligned}
\end{equation}
The gradient of the KLU loss with respect to $z_i$ is,
\begin{equation}\label{equ:KLUgradient}
\begin{aligned}
\frac{\partial \mathcal{L}_{\mathrm{KLU}}}{\partial z_i} &=-\sum_{j=1}^{C}\frac{\partial \left(\frac{1}{C}\log\left(C\cdot\hat{y}_{j}\right)\right)}{\partial \hat{y}_j}\cdot \frac{\partial \hat{y}_j}{\partial z_i} = \hat{y}_{i}-\frac{1}{C}.
\end{aligned}
\end{equation}

The magnitude of gradient decreases to zero when $\hat{y}_{i}\rightarrow \frac{1}{C}$, which also provides better convergence performance than the CE loss. We provide a detailed derivation of these gradients in the Appendix~\ref{a:soft}$\sim$\ref{a:klu}.

\textbf{The Difference between ICE and KLU.} From Equation~\eqref{equ:ICE} and Equation~\eqref{equ:KLU}, we can see that the ICE loss requires labelled data while the KLU loss does not. More specifically, the computation of the ICE loss involves both the data sample $x_i$ and the classification label $y_i$ from the restricted domain, while the computation of the KLU loss involves only the data sample $x_i$ from the restricted domain. This means that KLU suppresses any classification tasks regarding data samples in the restricted domain, but ICE suppresses only a certain classification task (specified by the classification label) in the restricted domain. For example, we want to suppress racial profiling task based on facial images. In this case, given facial images of data samples and corresponding labels, ICE suppresses only the racial classification task, and can preserve the performance on other classification tasks, \textit{e.g.}, gender identification. In contrast, KLU only depends on the data samples, and will suppress all classification tasks based on facial images. In this sense, KLU provides a more indiscriminate suppression compared with ICE. We evaluate the effectiveness and stability of these two loss functions in \S\ref{sucsec:stability}.

\subsubsection{Generation} For diffusion models, we use the MSE loss $\mathcal{L}_\mathrm{MSE}$ for the normal training reinforcement loss $\mathcal{L}_\beta$. Similarly, using $\mathcal{L}_{\mathrm{MSE}}$ for $\mathcal{L}_\alpha$ will also be problematic,
\begin{equation}\label{equ:MSEgradient}
\begin{aligned}
\frac{\partial \mathcal{L}_{\mathrm{MSE}}}{\partial \hat{\epsilon}_i} &=\frac{\partial \left(\sum_{j}\left(\epsilon_j-\hat{\epsilon}_j\right)^2\right)}{\partial \hat{\epsilon}_i} = 2 \left(\hat{\epsilon}_i-\epsilon_i\right).
\end{aligned}
\end{equation}

Therefore, we design a \textbf{Denial of Service (DoS)} loss,
\begin{equation}\label{equ:DoS}
\begin{aligned}
\mathcal{L}_{\mathrm{DoS}}(f)=\frac{1}{|\mathcal{X}|}\sum_i\|f(x_{ti}, t_i)\|^2.
\end{aligned}
\end{equation}
The gradient of the DoS loss with respect to $\hat{\epsilon}_i$ is $\frac{\partial \mathcal{L}_{\mathrm{DoS}}}{\partial \hat{\epsilon}_i} =\frac{\partial \left(\sum_{j}\left(\hat{\epsilon}_j\right)^2\right)}{\partial \hat{\epsilon}_i} = 2\hat{\epsilon}_i$.
The magnitude of this gradient decreases to zeros when $\hat{\epsilon}_i\rightarrow 0$, which provides better convergence than the MSE loss. Therefore, we use $\mathcal{L}_\mathrm{DoS}$ as the loss for $\mathcal{L}_\alpha$.

\section{Evaluation}\label{sec:evaluation}
\begin{figure}[tt]
    \centering
\setlength{\abovecaptionskip}{4pt}
\setlength{\belowcaptionskip}{-5pt}

\includegraphics[width=3.3in, trim=30 5 35 5, clip]{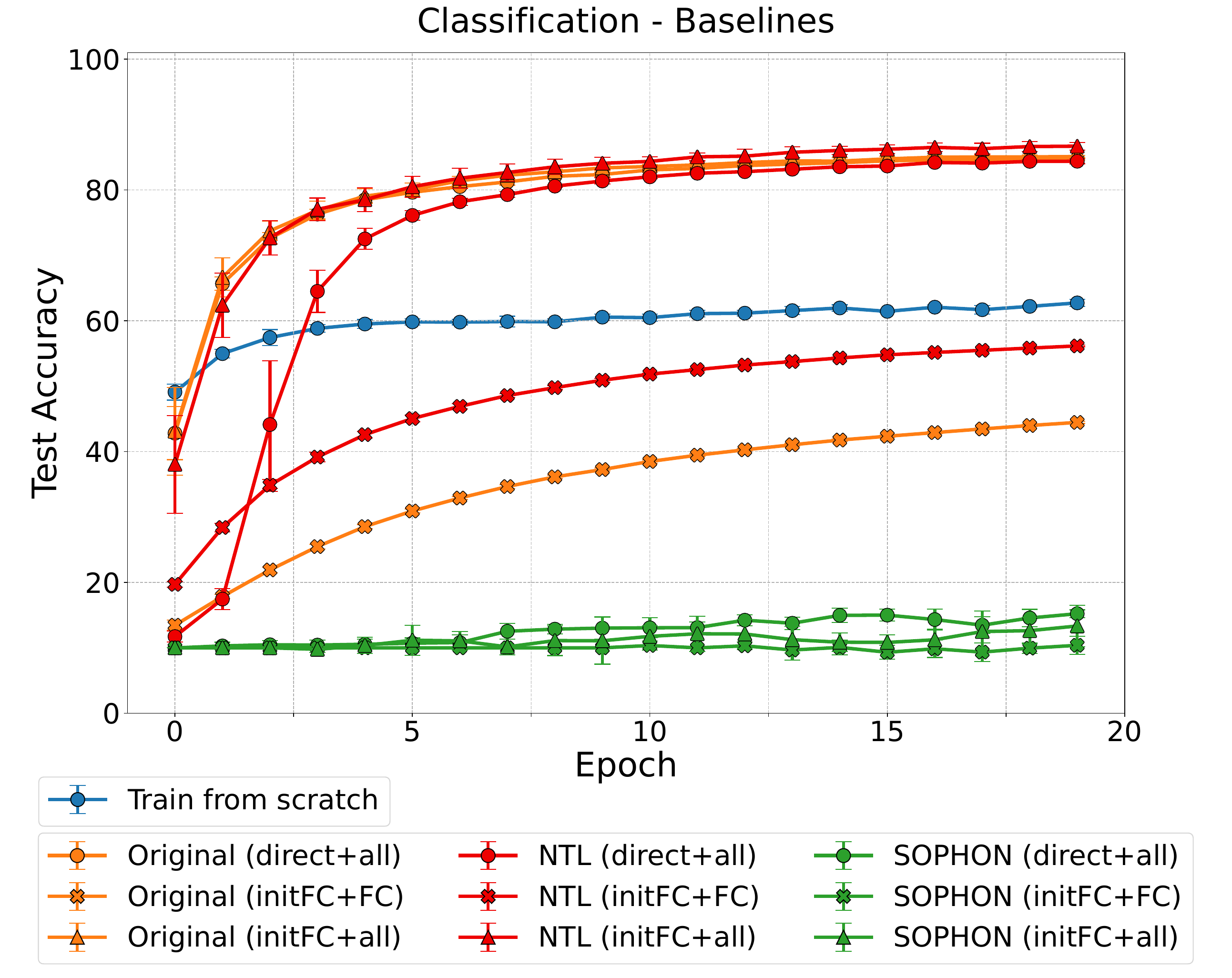}

\caption{Effectiveness of \sys compared with three baselines. Fine-tuning the original or the NTL model can achieve a high accuracy, leading to model misuse. All of the three methods of fine-tuning the \sys model yield poorer performances than training the model from scratch.}\label{fig:baselines}
\end{figure}

\begin{figure}[tt]
    \centering
\setlength{\abovecaptionskip}{4pt}
\setlength{\belowcaptionskip}{-5pt}

\includegraphics[width=3.3in, trim=15 10 15 5, clip]{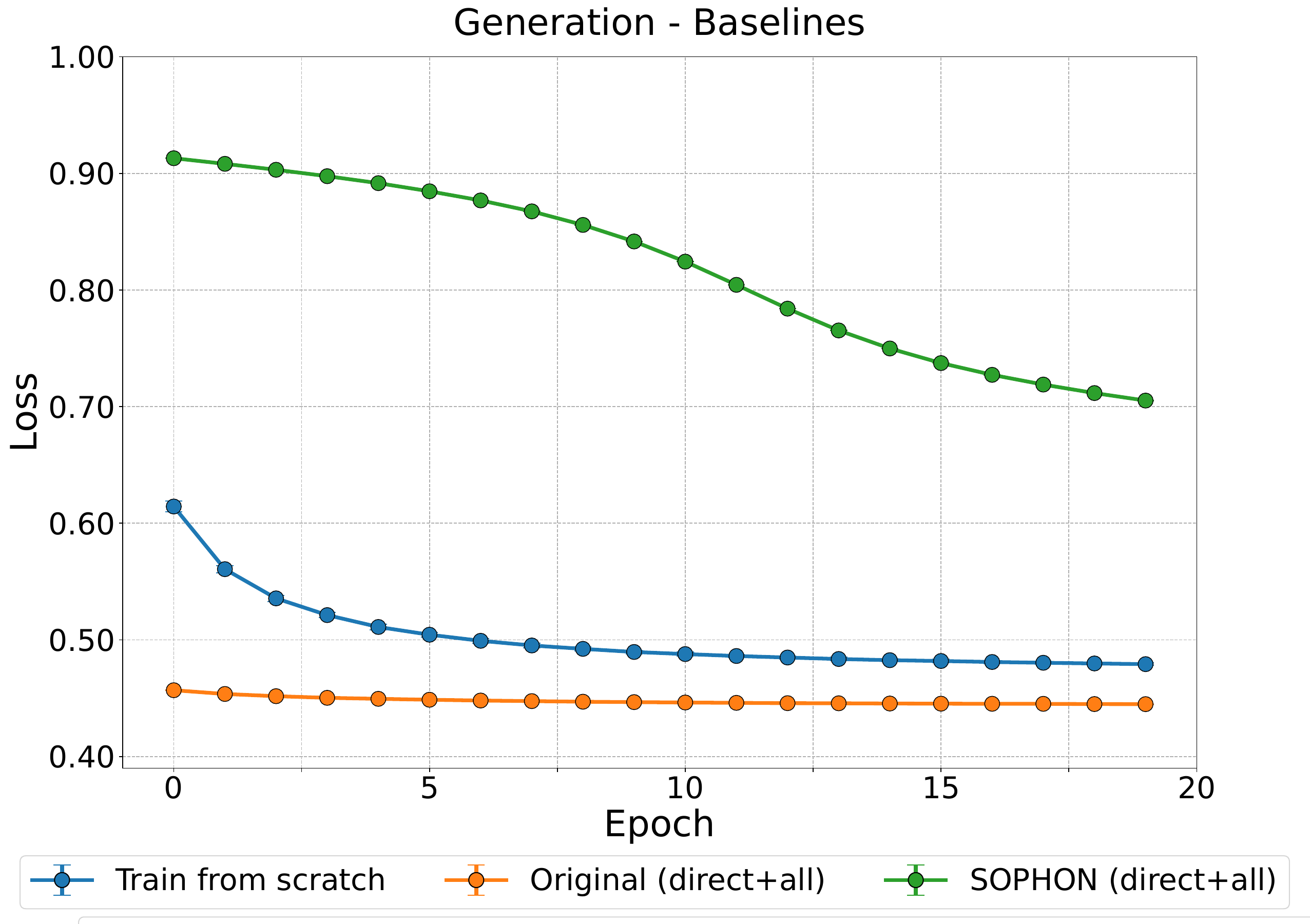}

\caption{Effectiveness of \sys compared with two baselines. Fine-tuning the original model can achieve low losses, leading to model misuse. Fine-tuning the \sys model yields higher losses than training the model from scratch.}\label{fig:baselines-gen}
\end{figure}

\begin{table*}\centering
\resizebox{\linewidth}{!}{
\begin{threeparttable}[t]
\setlength{\abovecaptionskip}{0pt}%
\setlength{\belowcaptionskip}{0pt}%
\caption{The overall effectiveness of \sys compared with three baselines (classification).}

\setlength{\tabcolsep}{3mm}{
\begin{tabular}{@{}lc|c|cccccc@{}}
\toprule
\multirow{2}{*}{Model}          & \multirow{2}{*}{\begin{tabular}[c]{@{}c@{}}Fine-tuning\\ Strategy\tnote{\textdagger}\end{tabular}} & Original ACC ($\uparrow$)         & \multicolumn{6}{c}{ACC in the Restricted Domain ($\downarrow$)}                                                 \\
                                &                                                                                 & Epoch 0               & Epoch 0               & Epoch 1      & Epoch 5      & Epoch 10     & Epoch 15     & Epoch 20     \\ \midrule
Train from scratch              & -                                                                               & -                     & -                     & \textbf{49.1}$\pm$1.2 & \textbf{59.5}$\pm$0.7 & \textbf{60.6}$\pm$0.5 & \textbf{62.0}$\pm$0.5 & \textbf{62.7}$\pm$0.5 \\ \midrule
\multirow{3}{*}{Original Model} & direct+all                                                                      & \multirow{3}{*}{99.6} & \multirow{3}{*}{9.3}  & \textbf{42.8}$\pm$4.0 & \textbf{78.5}$\pm$0.4 & \textbf{82.4}$\pm$0.3 & \textbf{84.2}$\pm$0.2 & \textbf{84.8}$\pm$0.2 \\
                                & initFC+FC                                                                       &                       &                       & \textbf{13.5}$\pm$0.8 & \textbf{28.5}$\pm$0.3 & \textbf{37.3}$\pm$0.2 & \textbf{41.8}$\pm$0.2 & \textbf{44.5}$\pm$0.2 \\
                                & initFC+all                                                                      &                       &                       & \textbf{43.1}$\pm$6.7 & \textbf{79.1}$\pm$1.1 & \textbf{83.4}$\pm$0.4 & \textbf{84.4}$\pm$0.5 & \textbf{85.1}$\pm$0.2 \\ \midrule
\multirow{3}{*}{NTL Model}      & direct+all                                                                      & \multirow{3}{*}{90.2} & \multirow{3}{*}{10.3} & \textbf{11.7}$\pm$0.9 & \textbf{72.5}$\pm$1.6 & \textbf{81.4}$\pm$0.5 & \textbf{83.6}$\pm$0.3 & \textbf{84.4}$\pm$0.4 \\
                                & initFC+FC                                                                       &                       &                       & \textbf{19.7}$\pm$0.4 & \textbf{42.6}$\pm$0.5 & \textbf{50.9}$\pm$0.3 & \textbf{54.3}$\pm$0.2 & \textbf{56.2}$\pm$0.1 \\
                                & initFC+all                                                                      &                       &                       & \textbf{38.1}$\pm$7.5 & \textbf{78.5}$\pm$1.8 & \textbf{84.1}$\pm$0.9 & \textbf{86.1}$\pm$0.6 & \textbf{86.7}$\pm$0.6 \\ \midrule
\multirow{3}{*}{\sys Model}     & direct+all                                                                      & \multirow{3}{*}{96.2} & \multirow{3}{*}{10.0} & \textbf{10.0}$\pm$0.0 & \textbf{10.5}$\pm$0.8 & \textbf{13.0}$\pm$1.7 & \textbf{15.0}$\pm$1.1 & \textbf{15.2}$\pm$0.6 \\
                                & initFC+FC                                                                       &                       &                       & \textbf{10.0}$\pm$0.0 & \textbf{10.0}$\pm$0.0 & \textbf{10.0}$\pm$0.0 & \textbf{10.1}$\pm$1.1 & \textbf{10.4}$\pm$1.4 \\
                                & initFC+all                                                                      &                       &                       & \textbf{10.0}$\pm$0.0 & \textbf{10.4}$\pm$1.2 & \textbf{11.1}$\pm$3.6 & \textbf{10.9}$\pm$1.4 & \textbf{13.3}$\pm$3.1 \\ \bottomrule
\end{tabular}
}

\begin{tablenotes}[flushleft]
\item[\textdagger] ``direct'': use the entire pre-trained model. ``initFC'': use the entire pre-trained model but randomly initialize the last FC layer. ``all'': fine-tune the whole model. ``FC'': only fine-tune the last FC layer.
\end{tablenotes}

\label{tab:overall}
\end{threeparttable}}
\end{table*}

\begin{table*}\centering
\resizebox{\linewidth}{!}{
\begin{threeparttable}[t]
\setlength{\abovecaptionskip}{0pt}%
\setlength{\belowcaptionskip}{0pt}%
\caption{The overall effectiveness of \sys compared with two baselines (generation).}

\setlength{\tabcolsep}{1.5mm}{
\begin{tabular}{@{}lc|c|cccccc@{}}
\toprule
\multirow{2}{*}{Model}          & \multirow{2}{*}{\begin{tabular}[c]{@{}c@{}}Fine-tuning\\ Strategy\tnote{\textdagger}\end{tabular}} & Original Loss ($\downarrow$)         & \multicolumn{6}{c}{Loss in the Restricted Domain ($\uparrow$)}                                                 \\
                                &                                                                                 & Epoch 0               & Epoch 0               & Epoch 1      & Epoch 5      & Epoch 10     & Epoch 15     & Epoch 20     \\ \midrule
Train from scratch              & -                                                                               & -                     & -                     & \textbf{0.614}$\pm$4.6e-3 & \textbf{0.511}$\pm$2.4e-3 & \textbf{0.490}$\pm$1.5e-3 & \textbf{0.483}$\pm$1.3e-3 & \textbf{0.479}$\pm$1.1e-4 \\ \midrule
Original Model & direct+all                                                                      & 0.478 & 0.477   & \textbf{0.457}$\pm$0.3e-4 & \textbf{0.449}$\pm$0.3e-4 & \textbf{0.447}$\pm$0.1e-4 & \textbf{0.445}$\pm$0.5e-4 & \textbf{0.445}$\pm$0.5e-4 \\
\sys Model     & direct+all                                                                      & 0.500 & 0.990   & \textbf{0.913}$\pm$0.9e-4 & \textbf{0.892}$\pm$1.8e-4 & \textbf{0.842}$\pm$1.1e-4 & \textbf{0.750}$\pm$0.2e-4 & \textbf{0.705}$\pm$0.7e-4 \\\bottomrule
\end{tabular}
}

\begin{tablenotes}[flushleft]
\item[\textdagger] The usual practice for fine-tuning generative models is to directly fine-tune the whole model.
\end{tablenotes}

\label{tab:overallgen}
\end{threeparttable}}
\end{table*}

\subsection{Setup}

\subsubsection{Prototype} We have implemented a prototype of \sys on the PyTorch \cite{DBLP:conf/nips/PaszkeGMLBCKLGA19} platform and processed the models according to Equation~\eqref{equ:obj2} and Algorithm~\ref{alg:sophon} using three NVIDIA A100 (80G) GPUs. We set the default \sys configurations as $\alpha=3\times 10^{-4}$, $\beta=5\times 10^{-4}$, $\mathrm{Iter}=800$, $K=50$, $\ell_\mathrm{FTS}=1$, and $\ell_\mathrm{NTR}=1$. One fine-tuning task per GPU is sampled in one FTS loop. We construct the tasks by randomly sampling the learning rates from $[10^{-6},10^{-5},10^{-4},10^{-3},10^{-2}]$ and the batch sizes from $[50,100,150,200,250]$, which cover the commonly-used range of learning rates and batch sizes. We always use the Adam optimizer in the fine-tuning simulation to simulate a powerful adversary.

The default fine-tuning strategy for evaluation is using the Momentum optimizer with a learning rate of $10^{-4}$, a batch size of 200, and a weight decay rate of $10^{-4}$ to directly fine-tune the whole model, unless specified otherwise. 


\subsubsection{Baselines}\label{subsec:baseline}
We compare \sys with three baselines to evaluate its effectiveness, \textit{i.e.}, 

\begin{itemize}
\setlength{\itemsep}{5pt}

    \item \textbf{Train from scratch (B1).} We train a model from scratch on the data sampled from the restricted domain as the first baseline. Note that, if the overhead of fine-tuning the pre-trained model is greater than training the model from scratch, then the adversary will have no incentive to fine-tune the pre-trained model. 
    \item \textbf{Fine-tune the original model (B2).} We fine-tune the original pre-trained model as our second baseline. Generally speaking, fine-tuning a pre-trained model can leverage the transferable knowledge and thus perform better than B1 in terms of the final performance and the time overhead. As mentioned in our design goal (\S\ref{subsec:goal}), we want the processed model to perform much worse than B2.
    \item \textbf{Fine-tune the NTL model (B3).} We reproduce the recent non-transferable learning~\cite{wang2022non} method as the third baseline for classification, for which it is specifically designed. NTL is proposed to degrade the model performance in the restricted domain. But the performance after fine-tuning is not considered. We demonstrate that NTL cannot resist fine-tuning via experiments in \S\ref{subsec:effectiveness}.
\end{itemize}

\subsubsection{Metrics} We adopt two metrics to evaluate the effectiveness of \sys, \textit{i.e.}, 

\begin{itemize}
\setlength{\itemsep}{5pt}

    \item \textbf{ACC-$k$}. The test accuracy at the $k$-th iteration of fine-tuning or training from scratch, which is a common metric for classification problems. 
    We want the protected model to have a low ACC in the restricted domain, even after being fine-tuned. 
    
    \item \textbf{MSE-$k$}. The mean squared error at the $k$-th iteration of fine-tuning or training from scratch, which is the training objective of the diffusion model. We want the protected model to have a high MSE in the restricted domain, even after being fine-tuned. 
    
    
    

    
    
\end{itemize}

We repeat each experiment five times with different random seeds and include the standard deviation for each outcome metric to mitigate the influence of randomness on the results.
The results are presented as $M \pm SD$, where $M$ denotes the mean value and $SD$ denotes the standard deviation.

\begin{figure*}[tt]
    \centering
\setlength{\abovecaptionskip}{0pt}
\setlength{\belowcaptionskip}{10pt}

\includegraphics[width=7.in, trim=155 50 155 50, clip]{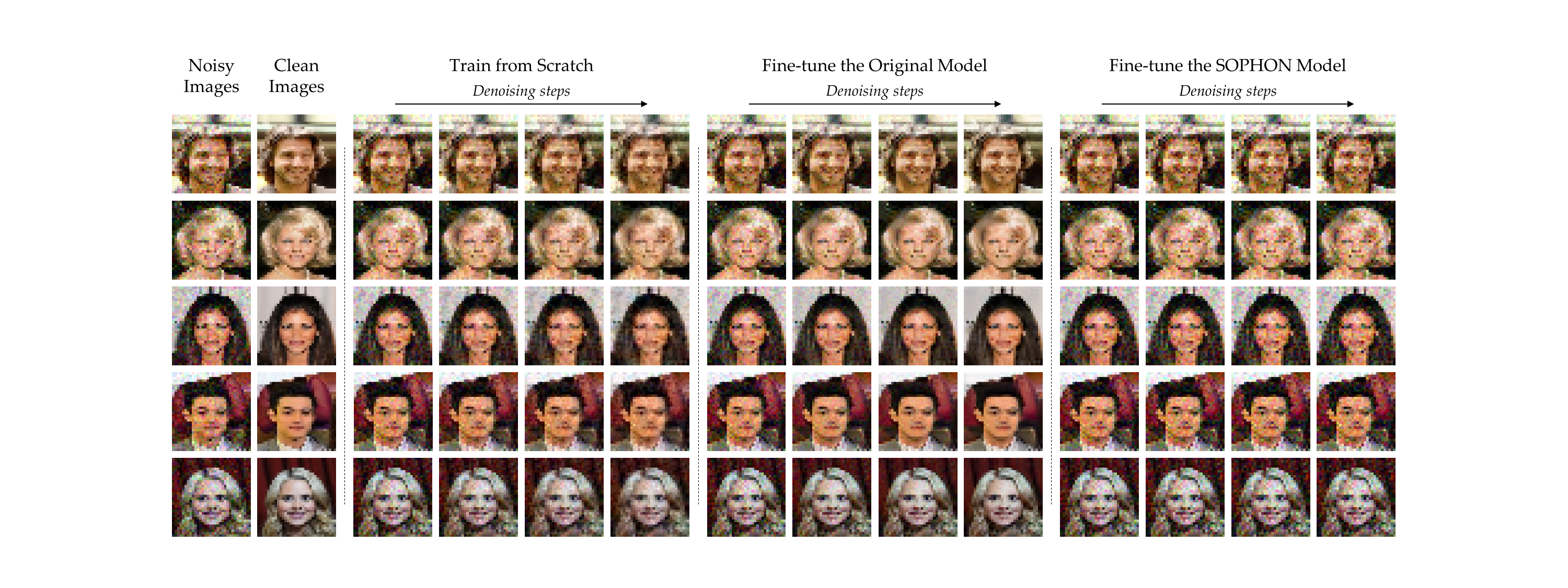}

\caption{Effectiveness of \sys compared with two baselines. B1 and B2 both perform well in the restricted domain (CelebA) in terms of the denoising ability. The \sys model cannot denoise images from the restricted domain, thus is protected.}\label{fig:baseline-gen2}
\end{figure*}

\begin{figure*}[tt]
    \centering
\setlength{\abovecaptionskip}{0pt}
\setlength{\belowcaptionskip}{10pt}

\includegraphics[width=7.in, trim=155 50 155 50, clip]{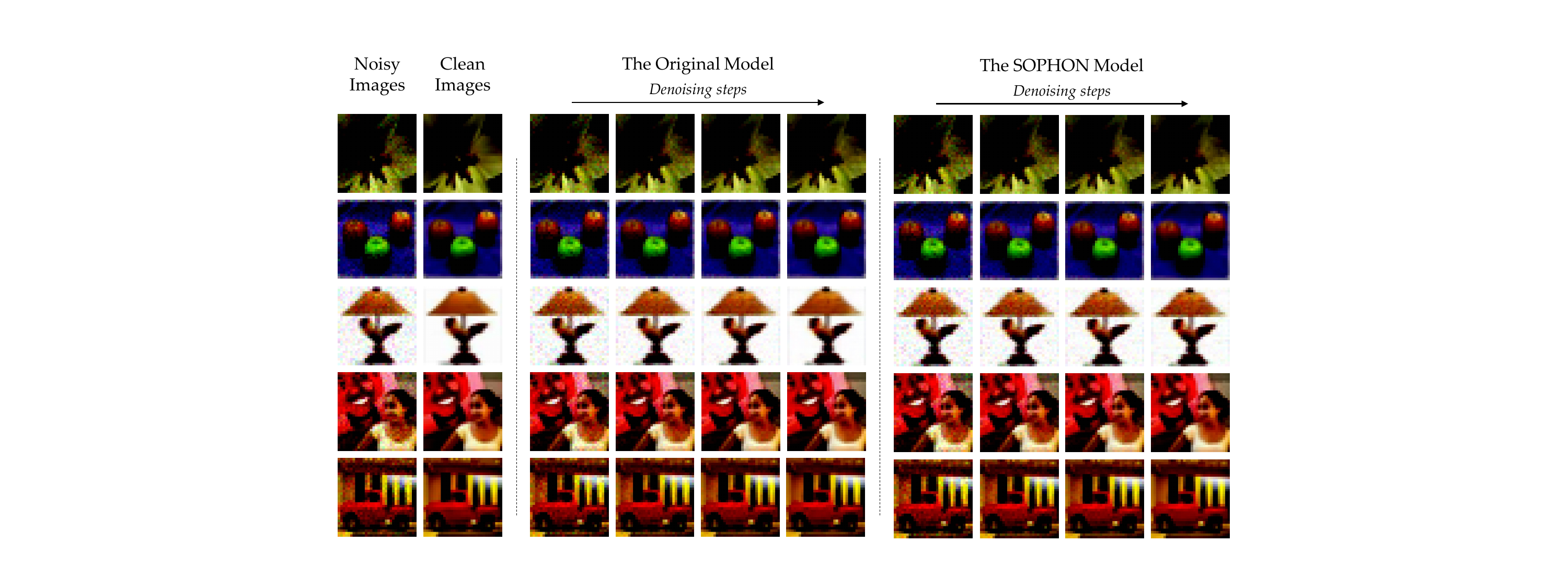}

\caption{Intactness of \sys. The \sys model performs well in the original domain (CIFAR-100) and has the denoising ability similar to the original model.}\label{fig:baseline-gen3}
\end{figure*}

\subsubsection{Datasets} Our experiments involve eight datasets, \textit{i.e.},

\begin{itemize}
\setlength{\itemsep}{5pt}

    \item \textbf{ImageNette}. ImageNette~\cite{imagenette} is a subset of 10 relatively simple classes from ImageNet.
    There are 12,894 training images and 500 test images. We process the images into 64$\times$64. Our pre-trained classification model is originally trained on this dataset.
    
    \item \textbf{CIFAR-10, CIFAR-100}. CIFAR-10~\cite{krizhevsky2009learning} consists of 32$\times$32 color images in 10 classes of animals and vehicles.
    There are 50,000 training images and 10,000 test images. CIFAR-100 is just like the CIFAR-10, except it has 100 classes. Our pre-trained generative model is originally trained on CIFAR-100.

    \item \textbf{CINIC}. CINIC-10~\cite{Darlow2018CINIC} consists of 32$\times$32 color images in 10 classes. It is an extension of CIFAR-10 via the addition of downsampled ImageNet images. There are 90,000 training images and 90,000 test images.
    
    \item \textbf{STL}.  STL-10~\cite{CoatesNL11} consists of 96$\times$96 color images in 10 classes of animals and vehicles like CIFAR-10. There are 5,000 training images and 8,000 test images.

    \item \textbf{MNIST}. MNIST~\cite{lecun1998mnist} consists of 28$\times$28 grayscale images in 10 classes (from ``0'' to ``9''). There are 60,000 training images and 10,000 test images.
    
    \item \textbf{SVHN}. SVHN~\cite{Yuval2011Reading} consists of 32$\times$32 color images in 10 classes (from ``0'' to ``9''). There are 73,257 training images and 26,032 test images.
    
    
    \item \textbf{CelebA}. CelebA~\cite{LiuLWT15} is a large-scale face dataset with $\sim$200,000 celebrity images of 218$\times$178. We split the dataset into 160,000 training images and 40,000 test images. CelebA is used as the restricted domain of the generative model in our experiments.
    
    \item \textbf{FFHQ}. FFHQ~\cite{KarrasLA19} consists of 1,024$\times$1,024 facial images with considerable variation in terms of age, ethnicity and image background. We use 8,000 images in FFHQ for training and 2,000 for testing. FFHQ is also 
    used as the restricted domain of the generative model in our experiments.
\end{itemize}

For those datasets that are used as the restricted domain, we split the training set of each dataset equally into two parts, one for \sys training or NTL (B3) training, the other one for the fine-tuning in the evaluation. This simulates the defender and the adversary obtaining different training data sampled from the same restricted domain.



    


    

\begin{figure*}[tt]
    \centering
\setlength{\abovecaptionskip}{0pt}
\setlength{\belowcaptionskip}{10pt}

\includegraphics[width=7.in, trim=0 5 0 5, clip]{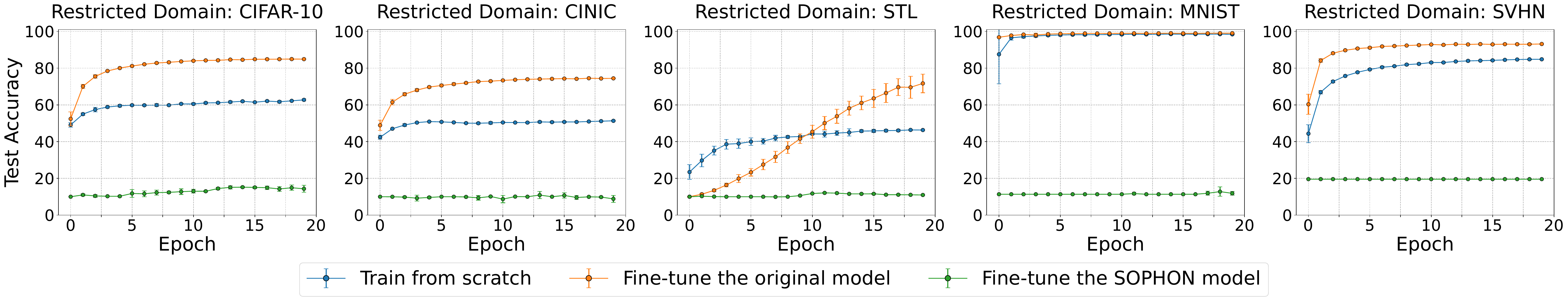}

\caption{Effectiveness of \sys against different restricted domains. Fine-tuning the \sys model for 20 epochs (5,000 iterations) cannot increase the test accuracy (\textit{e.g.}, from 10.0\% to 14.3\% for CIFAR-10), which means \sys can restrict the applicability of models in different domains.}\label{fig:domain}
\end{figure*}

\begin{figure*}[tt]
    \centering
\setlength{\abovecaptionskip}{0pt}
\setlength{\belowcaptionskip}{10pt}

\includegraphics[width=7.in, trim=0 5 0 5, clip]{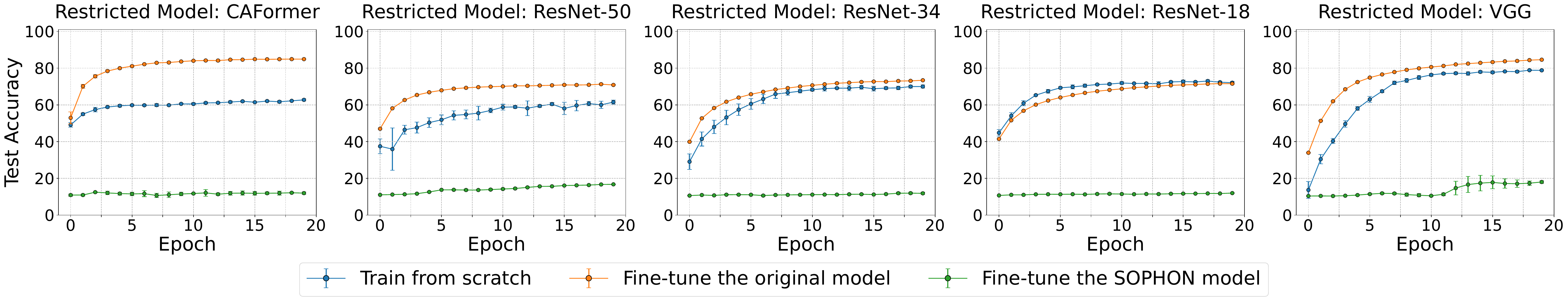}

\caption{Effectiveness of \sys on different model architectures. Fine-tuning the \sys models of different architectures for 20 epochs (5,000 iterations) only increases the test accuracy from 10.0\% to 18.0\% (VGG, the worst case), which means \sys achieving satisfying performances on model architectures of different model sizes.}\label{fig:model}
\end{figure*}

\subsection{Overall Effectiveness}\label{subsec:effectiveness}

\subsubsection{Classification}\label{subsec:effectiveness:classification}
In this part, we evaluate the overall effectiveness of \sys on classification tasks. We pre-train the CAFormer model on ImageNette and try to restrict its application on CIFAR-10. We consider three kinds of fine-tuning strategies, \textit{i.e.}, directly fine-tuning all the parameters (``direct+all''), randomly initializing the last FC layer and fine-tuning it (``initFC+FC''), and randomly initializing the last FC layer and fine-tuning all the parameters (``initFC+all''). We fine-tune the baseline models and our model for a total of $20$ epochs ($20\times250=5,000$ iterations in total).

For the classification task, we compare our method with the three baselines mentioned in \S\ref{subsec:baseline}. We test the accuracy of the models every epoch and present the results in Figure~\ref{fig:baselines} and Table~\ref{tab:overall}. We can see that by training the model from scratch, the model finally reaches an ACC of 62.7\% at 5,000 iterations. Fine-tuning the original model and the NTL model using the ``direct+all'' or ``initFC+all'' strategies can achieve much higher ACCs of 84.4\%$\sim$86.7\%, which shows the superior benefit brought by pre-training. This also verifies that existing NTL methods cannot resist supervised fine-tuning since they actually do not include a fine-tune-related objective in their optimization problem. Fine-tuning the models using the ``initFC+FC'' strategy can only obtain performances close to training from scratch. This is aligned with the findings of a previous study~\cite{YosinskiCBL14}. Compared with the baselines, the \sys model resists fine-tuning and finally yields a far lower ACC of 10.4\%$\sim$15.2\%, which is comparable to a random guess. The results show the effectiveness of our method that it turns a pre-trained model into a non-fine-tunable model against a specific restricted domain, while preserving a high ACC in the original domain of 96.2\%. In other words, fine-tuning the \sys models is much harder than fine-tuning the original models and even harder than training from scratch, which means the adversary will have no incentive to abuse the protected pre-trained models.

\subsubsection{Generation}\label{subsec:effectiveness:generation}
In this part, we evaluate the effectiveness of \sys on generation tasks. We pre-train the diffusion model on CIFAR-100 and try to restrict its application on CelebA. In other words, we do not want the model to be used for generating fake faces. We adopt the ``direct+all'' fine-tuning strategy since the diffusion model is typically a U-Net model that lacks an explicit component to randomly initialize.

The results comparing \sys with B1 and B2 are shown in Figure~\ref{fig:baselines-gen} and Table~\ref{tab:overallgen}. Directly fine-tuning the unprotected model can easily yield a very low MSE, \textit{e.g.}, 0.457 in one epoch. By training from scratch, a relatively satisfactory result is obtained in 10 epochs with an MSE of 0.490, a bit worse than B1. In contrast, fine-tuning the \sys model only achieves an MSE of 0.705 in even 20 epochs, which shows great effectiveness of \sys on generation tasks.

We also present the images generated by the two baselines and our method in Figure~\ref{fig:baseline-gen2} and Figure~\ref{fig:baseline-gen3}. The images are added noises in the forward process, and are denoised by the models. From Figure~\ref{fig:baseline-gen2}, we can see that fine-tuning the original model in the restricted domain yields very good performance in the restricted domain. Training the model from scratch in the restricted domain also achieve relatively good results even though the results are not as good as fine-tuning the original model. In comparison, the diffusion model fine-tuned from the \sys model almost has no ability to denoise facial images, verifying the non-fine-tunability in the restricted domain. Note that, Figure~\ref{fig:baseline-gen3} shows that the \sys model actually performs well in the original domain in terms of the denoising quality, which means \sys achieves intactness. 

\begin{figure*}[tt]
    \centering
\setlength{\abovecaptionskip}{0pt}
\setlength{\belowcaptionskip}{10pt}

\includegraphics[width=7.in, trim=0 5 0 5, clip]{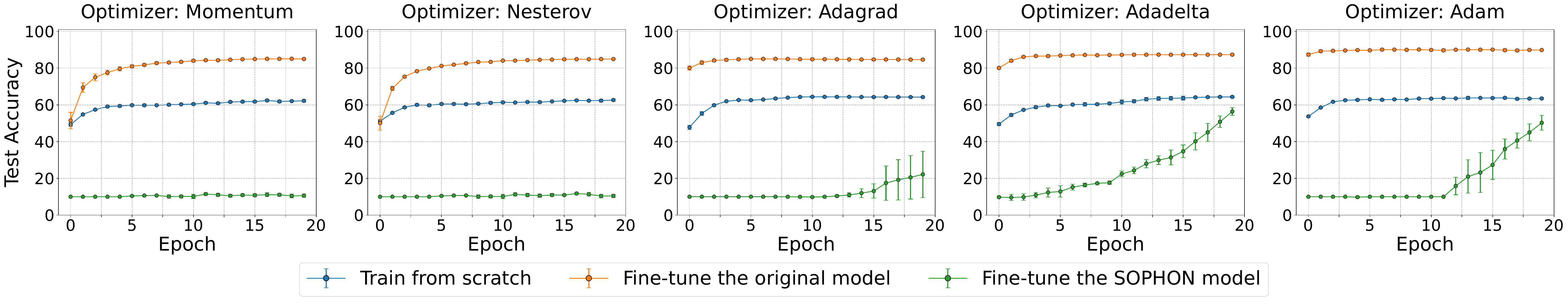}

\caption{Effectiveness of \sys against different optimizers. Momentum or Nesterov does not increase the accuracy. Adagrad, Adadelta, and Adam, may increase the accuracy in the restricted domain, but \sys still renders the model more challenging to train compared to training from scratch, greatly slowing down the convergence. The results verify that although we only use Adam in the NFT loops, the resistance generalizes well to other optimizers.}\label{fig:optimizer}
\end{figure*}

\begin{figure*}[tt]
    \centering
\setlength{\abovecaptionskip}{0pt}
\setlength{\belowcaptionskip}{10pt}

\includegraphics[width=7.in, trim=0 5 0 5, clip]{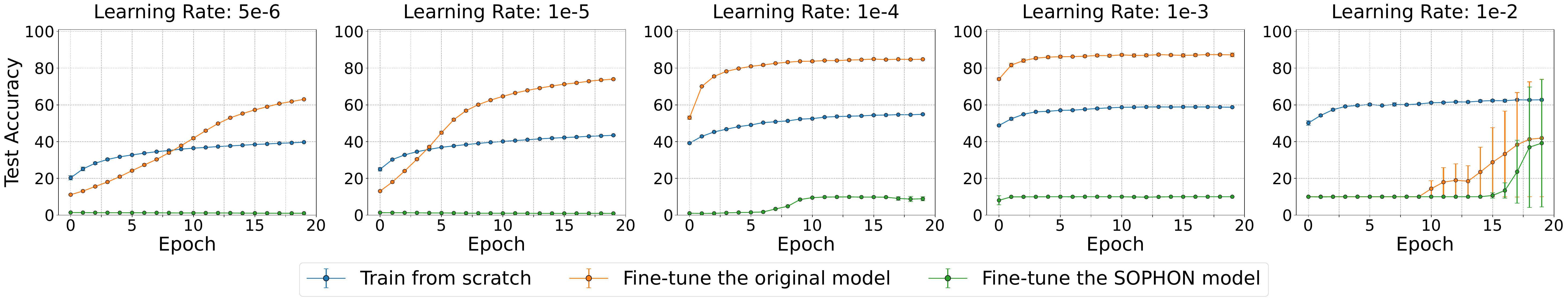}

\caption{Effectiveness of \sys against different learning rates. The values $10^{-4}$ and $10^{-3}$ are both suitable as learning rates. Even under such suitable learning rates, the \sys model still resists fine-tuning, showing robustness against different learning rates.}\label{fig:lr}
\end{figure*}

\subsection{Restricted Domain}
In this part, we evaluate the effectiveness of \sys against different restricted domains. For classification, we pre-train the CAFormer model on ImageNette and try to restrict its application on object classification datasets (CIFAR-10, CINIC, and STL) and digit classification datasets (MNIST and SVHN). For generation, we pre-train the diffusion model on CIFAR-100 and restrict its application on two image datasets (CelebA and FFHQ).

We compare our method with the B1 and B2 baselines and present the classification results in Figure~\ref{fig:domain} and Table~\ref{tab:domain} in Appendix. The three models trained from scratch on CIFAR-10, CINIC, and STL only achieve ACCs of 62.7\%, 51.3\%, and 46.3\%, respectively. These results align with the expected difficulty levels of the three tasks. Fine-tuning the original models achieves 84.9\%, 74.4\%, and 71.7\%, much higher than training from scratch. Since MNIST and SVHN are relatively easier, both B1 and B2 of MNIST and SVHN achieve high ACCs. Note that B2 still outperforms B1, which is the gain from pre-training. Compared with the baselines, fine-tuning the \sys models only results in performance comparable to random guessing, \textit{i.e.}, 8.8\%$\sim$19.6\%. For example, fine-tuning the \sys model on CIFAR-10 for 20 epochs only increases the ACC from 10.0\% to 14.3\%. As we can see, \sys can restrict the applicability of models in different domains. Meanwhile, the \sys models still maintain good performance in the original domain, \textit{i.e.}, 96.2\%$\sim$97.8\%. Similar conclusions can be inferred from the results of generation tasks depicted in Figure~\ref{fig:baselines-gen} and Figure~\ref{fig:baselines-gen4} in Appendix.

\subsection{Model Architecture}

In this part, we evaluate the effectiveness of \sys on different model architectures. We pre-train five models (CAFormer, ResNet-50, ResNet-34, ResNet-18, and VGG) on ImageNette and try to restrict their application on CIFAR-10.

As shown in Figure~\ref{fig:model} and Table~\ref{tab:model} in Appendix, we can still observe that the \sys models resist fine-tuning and can only result in low ACCs. Note that for five model architectures, \sys achieves ACCs in the original domain of 96.2\%, 95.6\%, 94.8\%, 95.0\%, and 91.2\% on CAFormer, ResNet-50, ResNet-34, ResNet-18, and VGG. This trend is almost in line with the number of parameters of the models, \textit{i.e.}, 53.9M, 23.5M, 21.3M, 11.2M and 9.6M. The results indicate that \sys achieves satisfying intactness on different model architectures. Given that maintaining performance in the original domain and degrading performance in the restricted domain are dual objectives, it is plausible that a more complex model, with an increased number of parameters, holds the potential to better achieve these dual objectives. This is particularly attainable in the current landscape, where large-scale pre-training is a prevailing trend.

\begin{figure*}[tt]
    \centering
\setlength{\abovecaptionskip}{0pt}
\setlength{\belowcaptionskip}{10pt}

\includegraphics[width=7.in, trim=0 5 0 5, clip]{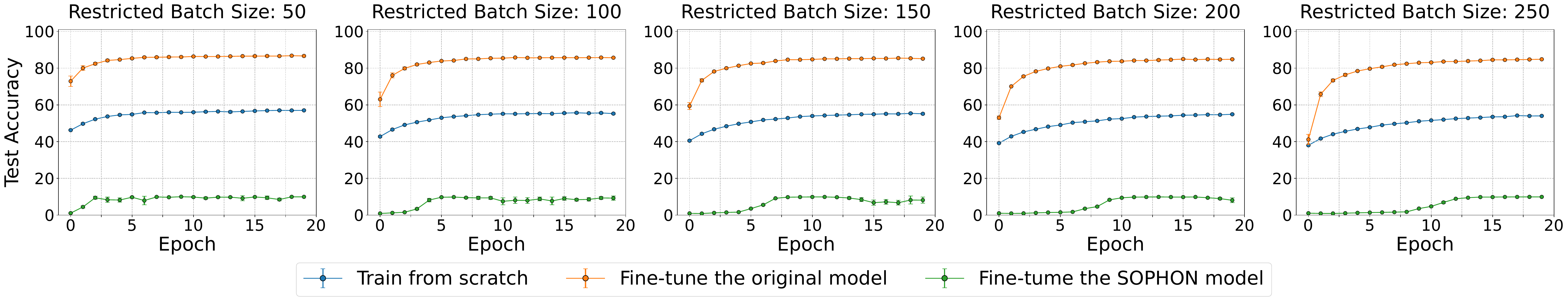}

\caption{Effectiveness of \sys against different batch sizes. Note that, although the \sys model is updated 20,000 times when the batch size is 50, the model is still stuck in the local optimum. It verifies the robustness of \sys against different batch sizes.}\label{fig:bs}
\end{figure*}

\subsection{Optimizer}

In this part, we evaluate the effectiveness of \sys against different optimizers. We pre-train the CAFormer model on ImageNette and try to restrict its application CIFAR-10. The models are fine-tuned by five different optimizers, \textit{i.e.}, Momentum, Nesterov, Adagrad, Adadelta, and Adam.

We present the results in Figure~\ref{fig:optimizer} and Table~\ref{tab:optimizer} in Appendix. The results show that fine-tuning the \sys models using the Momentum or the Nesterov optimizer does not increase the ACC, which remains almost unchanged at around 10\%. Leveraging optimizers that adaptively adjust the learning rate, \textit{i.e.}, Adagrad, Adadelta, and Adam, may increase the performance of \sys models in the restricted domain. But \sys still renders the model more challenging to train than training from scratch, greatly slowing down the convergence. For example, training the model from scratch using Adadelta achieves 59.7\% in just 5 epochs. In contrast, fine-tuning the \sys using the same Adadelta takes 20 epochs to achieve 56.4\%. The results verify that although we only use Adam in the fine-tuning suppression loops, the resistance generalizes well to other optimizers.

We observe that the accuracy of fine-tuning the protected models using Adagrad, Adadelta, and Adam shows an upward trend in the later stages of the training process. To further investigate, we extend the training by an additional 30 epochs until the training processes converged. The final ACCs of fine-tuning the protected models using Adagrad, Adadelta, and Adam are 49.8\%$\pm$32.38\%, 78.8\%$\pm$0.62\%, and 75.6\%$\pm$1.58\%, respectively. It is noteworthy that fine-tuning the protected models still yields significantly lower performance compared to fine-tuning the original models, albeit slightly better than training from scratch. This suggests that if the adversary persists in fine-tuning the protected models, they can achieve a 10.1\% to 13.4\% improvement compared with training from scratch, albeit at the cost of expending 10 times the effort (40$\sim$50 epochs versus 3$\sim$4 epochs).

\subsection{Learning Rate}
In this part, we evaluate the effectiveness of \sys against different learning rates, because learning rates may impact whether a model can avoid local optima. 
We pre-train the CAFormer model on ImageNette and try to restrict its application CIFAR-10. 
The models are fine-tuned using five different learning rates, \textit{i.e.}, $5\times10^{-6}, 10^{-5}, 10^{-4}, 10^{-3}$, and $10^{-2}$. We do not evaluate $10^{-6}$ because it is too small for efficient model training.

The results are shown in Figure~\ref{fig:lr} and Table~\ref{tab:lr} in Appendix. We can tell from the results that $5\times10^{-6}$ and $10^{-5}$ result in a slow convergence, due to they are still too small. Conversely, $10^{-2}$ is commonly too large for fine-tuning because fine-tuning typically does not change the parameters too much. Thus, fine-tuning either the original model or the \sys model cannot reach a satisfying solution in this case. The values $10^{-4}$ and $10^{-3}$ are both suitable as learning rates, as they lead to relatively fast convergence when training from scratch and fine-tuning the original model. Even under such suitable learning rates, the \sys model still resists fine-tuning, showing robustness against different learning rates. This could be attributed to our inclusion of tasks with varying learning rates in the fine-tuning simulation.







\subsection{Batch Size}
In this part, we evaluate the effectiveness of \sys against different batch sizes because batch sizes impact the accuracy of gradient computation. We pre-train the CAFormer model on ImageNette and try to restrict its application CIFAR-10. The models are fine-tuned using five different batch sizes, \textit{i.e.}, 50, 100, 150, 200, and 250. Given that the batch size impacts the number of iterations in one epoch, the total iterations for these five settings over 20 epochs vary from 20,000 to 4,000.

The results in Figure~\ref{fig:bs} and Table~\ref{tab:batchsize} show that \sys is able to resist fine-tuning under these five batch size settings. Note that, although the \sys model is updated 20,000 times when the batch size is 50, the model is still stuck in the local optimum. It verifies the robustness of \sys against different batch sizes. 
This could also be attributed to our inclusion of tasks with varying batch sizes in the fine-tuning simulation.

\subsection{Training Stability}\label{sucsec:stability}
In this part, we verify the stability of our proposed three loss functions, the ICE loss, the KLU loss and the DoS loss.

\begin{figure}[tt]
    \centering
\setlength{\abovecaptionskip}{0pt}
\setlength{\belowcaptionskip}{10pt}

\subfigure{
    \includegraphics[width=2.3in, trim=0 5 1100 5, clip]{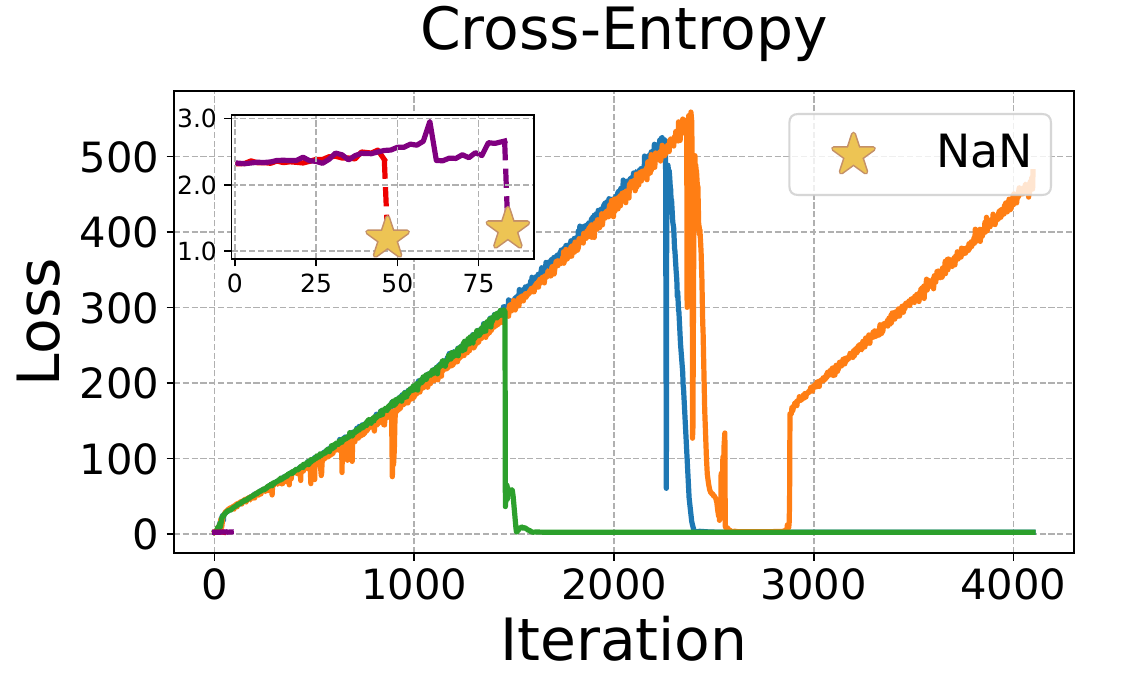}
    }
\\
\subfigure{
    \includegraphics[width=2.3in, trim=560 5 540 5, clip]{figures/loss_functions.pdf}
    }
\\
\subfigure{
    \includegraphics[width=2.3in, trim=1120 5 -20 5, clip]{figures/loss_functions.pdf}
    }

\caption{Training stability comparison of three loss functions for classification. The CE loss easily fails due to abrupt changes and divergence. The ICE loss and the KLU loss exhibit stability over 4,000 iterations, with the fluctuations in the loss decreasing.}\label{fig:clsloss}
\end{figure}

As a baseline for classification model training, we present five failure cases encountered during the optimization process using the CE loss, as shown in Figure~\ref{fig:clsloss}. In some cases, the loss becomes not-a-number (NaN) values during training. In other cases, the loss exhibits abrupt changes and divergence as it increases. It leads to underfitting and poor performance in the optimization process. On the contrary, the ICE loss and the KLU loss exhibit stability over 4,000 iterations, with the fluctuations in the loss decreasing. In our experiments, we use the two loss functions in our optimization over 200 times, no instability in training has been observed.

\begin{figure}[tt]
    \centering
\setlength{\abovecaptionskip}{0pt}
\setlength{\belowcaptionskip}{10pt}

\subfigure{
    \includegraphics[width=2.3in, trim=0 5 650 5, clip]{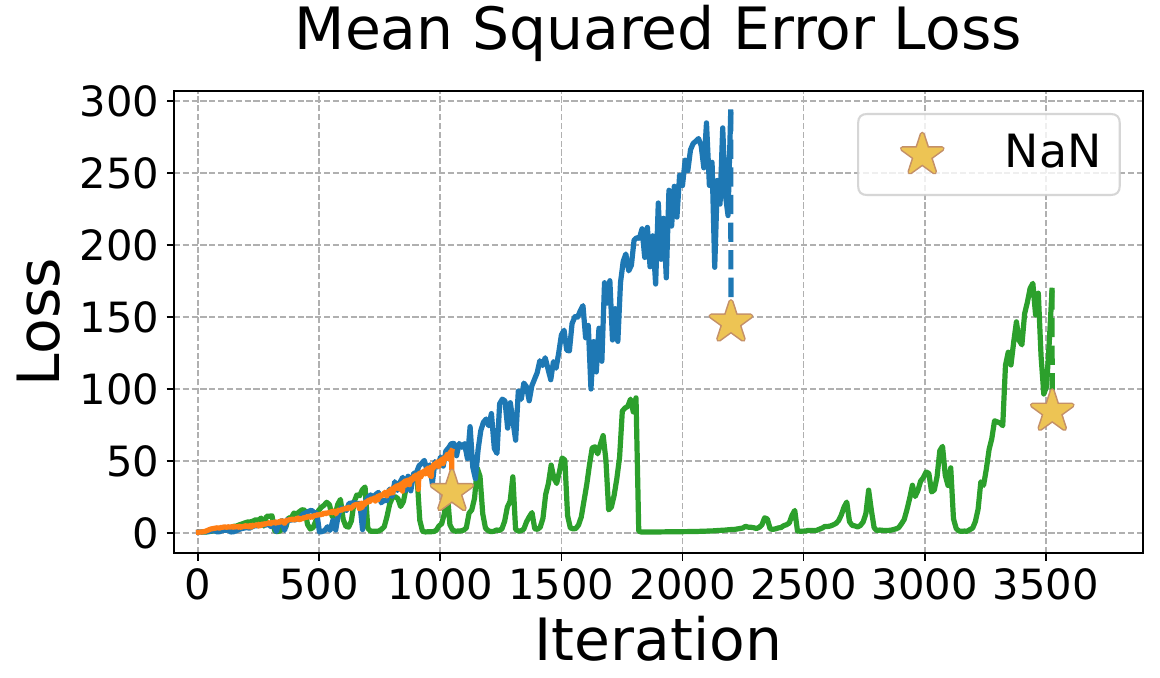}
    }
\\
\subfigure{
    \includegraphics[width=2.3in, trim=650 5 0 5, clip]{figures/stability_generation.pdf}
    }

\caption{Training stability comparison of two loss functions for generation. The MSE loss easily fails due to abrupt changes and divergence. The DoS loss exhibits stability over 5,000 iterations, facilitating easier and faster convergence during optimization.}\label{fig:genloss}
\end{figure}

As a baseline for generative model training, we present three failure cases encountered during the optimization process using the MSE loss, as shown in Figure~\ref{fig:genloss}. As we can see, the MSE loss also results in NaN values during training. And the green line in Figure~\ref{fig:genloss} exhibits several abrupt changes to zero, rendering the optimization inefficient and ineffective. Conversely, the DoS loss we design demonstrates stability throughout the training process, facilitating easier and faster convergence during optimization.

\section{Related Work}\label{sec:relatedwork}

\subsection{Non-Transferable Learning}
Non-transferable learning (NTL) aims to degrade the performance of deep learning models in the restricted domain, which is partially aligned with our work. Wang~\textit{et~al}.~\cite{wang2022non} proposed the first NTL method by decreasing the model performance in the restricted domain with labeled data and increasing the distance between features from two domains with the maximum mean discrepancy (MMD) loss.
Following the concept of NTL, Wang~\textit{et~al.}~\cite{wang2023model} focused on image classification and proposed a compact un-transferable isolation (CUTI)-domain method
to achieve a compacter generalization bound of the model. Zeng~\textit{et~al}.~\cite{zeng2022unsupervised} further extended the idea to natural language processing tasks. They leveraged a domain classifier loss and an MMD loss to eliminate the need for labeled data from the restricted domain. They also introduced a secret key component for recovering access to the restricted domain. Wang~\textit{et~al}.~\cite{wang2023domain} utilized distributionally robust optimization (DRO) framework to characterize the domains close to the source domain and degrade the model performance in them. In this way, they achieved better performance than NTL when no data is available from the restricted domain. 

Compared with existing NTL work, we not only degrade the model performance in the restricted domain but also resist fine-tuning the model in the restricted domain. We have confirmed via experiments that existing NTL methods cannot resist fine-tuning (cf.~\S\ref{subsec:effectiveness:classification}), probably because they are not robust against supervised model update. In addition, they only focused on classification tasks, while \sys can also be applied to generation (cf.~\S\ref{subsec:effectiveness:generation}).



\subsection{Meta Learning}
Meta learning, or learning-to-learn, aims to improve the learning algorithm itself, given the experience of multiple learning episodes. Model-agnostic meta-learning (MAML)~\cite{finn2017model,finn2018probabilistic,finn2019online} is a famous example to learn a model initialization that converges fast on unseen tasks and thus enables few-shot learning. Then a series of follow-up work was proposed to reduce the number of parameters to meta-learn. Lee~\textit{et~al}.~\cite{lee2018gradient} performed MAML in a subspace of the whole parameter space by freezing a portion of the neurons. Qiao~\textit{et~al}.~\cite{qiao2018few} and  Rusu~\textit{et~al}.~\cite{rusu2019meta} created a low-dimensional latent space for model parameters by learning a parameter generator. Sun~\textit{et~al}.~\cite{sun2019meta} leveraged large-scale trained DNN and only meta-learned the scaling and shifting functions of model parameters to further speed up convergence and avoid over-fitting. 

In comparison with meta learning, the objectives of non-fine-tunable learning are somewhat the opposite, \textit{i.e.}, to train the model to not learn specific tasks while preserving unchanged performance on the original task. We face unique challenges such as negative impact in the original domain, susceptibility to adaptive fine-tuning, and training instability. Thus, directly \textit{inverting} existing methods cannot fulfill our goals.


\section{Discussion \& Future Work}\label{sec:discussion}
In this section, we discuss the limitations and the future work of \sys.


\textit{More deep learning tasks.} In this paper, we have demonstrated the effectiveness of \sys in two specific tasks: image classification and image generation. While \sys technically holds potential for application in other domains, generalizing the results presented in this paper remains unclear. Therefore, we advocate for the application of \sys to pre-trained models in various domains, such as audio processing, natural language processing, tabular data analysis, and multimodal tasks. By extending \sys to these domains, we may unlock its potential for enhancing the controllability of models across diverse areas of machine learning and artificial intelligence, a direction we envision as our future work.




\textit{More domain-adaptation techniques.}
In this paper, we have demonstrated the effectiveness of \sys against several commonly used domain adaptation techniques. We have designed a fine-tuning simulation module in \sys to enhance its resistance to various domain adaptation techniques by incorporating them into $\Phi$. However, with the emergence of sophisticated domain adaptation techniques, such as LoRA/Adaptor in NLP, the effectiveness of \sys against these techniques remains unknown. We have conducted preliminary experiments of fine-tuning CAFormer with LoRA and found \sys to remain effective. We advocate for a more extensive and comprehensive evaluation on various domain adaptation techniques.

\textit{More efficient and accurate algorithms.} \sys achieves non-fine-tunable learning by simulating fine-tuning process in the optimization, which involves second-order terms and requires intensive computation. To reduce overhead, we use the first-order approximation in fine-tuning suppression, but the algorithm may be less accurate in this way. Designing an approximation algorithm that reduces the amount of computation while maintaining a satisfactory accuracy is a promising area for future research.


\section{Conclusion}\label{sec:conclusion}
In this paper, we introduce a pioneering learning paradigm, termed non-fine-tunable learning, aimed at preventing pre-trained models from being fine-tuned for restricted tasks. We present the problem formulation, design and evaluation of \sys, a framework designed to achieve non-fine-tunable learning, including an effective algorithm to resist fine-tuning and novel loss functions for training stability. Our experiments have validated the efficacy of \sys in resisting fine-tuning across various fine-tuning settings.



\section*{Acknowledgments}
We sincerely thank our Shepherd and all the anonymous reviewers for their valuable comments. This work is supported by China NSFC Grant 61925109 and Ant Group. Yanjiao Chen is the corresponding author.

\clearpage
\bibliographystyle{plain}
\bibliography{mybib}

\appendices
\section{Derivatives in this Paper}
\subsection{Derivatives of \texttt{Softmax}}\label{a:soft}
The output of a classifier can be represented as,
\begin{equation}\small
\begin{aligned}
 &\left(\hat{y}_1, \hat{y}_2, \cdots, \hat{y}_C\right) = \sigma\left(z_1, z_2, \cdots, z_C\right),
\end{aligned}
\end{equation}
where $z_i$ is the $i$-th logit, $\hat{y}_i = \exp(z_i)/\sum_{j=1}^C\exp(z_j)$ is the predicted probability of the $i$-th label, $C$ is the total number of classes. When $i\ne j$,

\begin{equation}\small
\begin{aligned}
    \frac{\partial \hat{y}_j}{\partial z_i} &=\frac{\partial \left(\exp(z_j)/\sum_{k=1}^C\exp(z_k)\right)}{\partial z_i},\\
&=\frac{-\exp(z_i)\cdot\exp(z_j)}{\left(\sum_{k=1}^C\exp(z_k)\right)^2},\\
&=-\hat{y}_i\hat{y}_j.
\end{aligned}
\end{equation}

When $i=j$,
\begin{equation}\small
\begin{aligned}
    \frac{\partial \hat{y}_i}{\partial z_i} &=\frac{\partial \left(\exp(z_i)/\sum_{k=1}^C\exp(z_k)\right)}{\partial z_i},\\
&=\frac{\exp(z_i)\left(\sum_{k=1}^C\exp(z_k)\right)-\exp(z_i)^2}{\left(\sum_{k=1}^C\exp(z_k)\right)^2},\\
&=\frac{\exp(z_i)}{\sum_{k=1}^C\exp(z_k)}-\left(\frac{\exp(z_i)}{\sum_{k=1}^C\exp(z_k)}\right)^2,\\
&=\hat{y}_i(1-\hat{y}_i).
\end{aligned}
\end{equation}

Then, we have the following partial derivatives,
\begin{equation}\label{equ:softmaxgradient}\small
\frac{\partial \hat{y}_j}{\partial z_i} = \left\{\begin{array}{rl}
       \hat{y}_i(1-\hat{y}_i), & i=j,\\
      -\hat{y}_i\hat{y}_j, & i\ne j.
\end{array}\right.
\end{equation}

\subsection{Derivatives of Cross-Entropy Loss}\label{a:ce}
We can derive the gradient of the cross-entropy loss with respect to $z_i$ by the chain rule,
\begin{equation}\small
\begin{aligned}
\frac{\partial \mathcal{L}_{\mathrm{CE}}}{\partial z_i} &=-\sum_{j=1}^{C}\frac{\partial \left(y_j\log \hat{y}_j\right)}{\partial \hat{y}_j}\cdot \frac{\partial \hat{y}_j}{\partial z_i}, \\
&=-\sum_{j=1}^{C}\frac{y_j}{\hat{y}_j}\cdot \frac{\partial \hat{y}_j}{\partial z_i}, \\
&=-\frac{y_i}{\hat{y}_i}\cdot\frac{\partial \hat{y}_i}{\partial z_i}-\sum_{j\ne i}\frac{y_j}{\hat{y}_j}\cdot \frac{\partial \hat{y}_j}{\partial z_i},\\
&\overset{\star}{=}-\frac{y_i}{\hat{y}_i}\cdot\hat{y}_i(1-\hat{y}_i)-\sum_{j\ne i}\frac{y_j}{\hat{y}_j}\cdot \left(-\hat{y}_i\hat{y}_j\right),\\
&= -y_i+\left(\sum_{j=1}^{C}y_j\right)\cdot \hat{y}_i,\\
&= \hat{y}_i - y_i,
\end{aligned}
\end{equation}
where Equation ($\star$) utilizes the results in Equation~\eqref{equ:softmaxgradient}.

\subsection{Derivatives of Inverse Cross-Entropy Loss}\label{a:ice}
Without loss of generality, we assume $y_1=1$ and $y_j=0~(j\ne 1)$. Then the gradient of the ICE loss with respect to $z_i$ is,
\begin{equation}\small
\begin{aligned}
\frac{\partial \mathcal{L}_{\mathrm{ICE}}}{\partial z_i} &=-\sum_{j=1}^{C}\frac{\partial \left(y_j\log \left(1-\hat{y}_{j}\right)\right)}{\partial \hat{y}_j}\cdot \frac{\partial \hat{y}_j}{\partial z_i},\\
&=\sum_{j=1}^{C}\frac{y_j}{1-\hat{y}_{j}}\cdot \frac{\partial \hat{y}_j}{\partial z_i},\\
&=\frac{y_i}{1-\hat{y}_{i}}\cdot \frac{\partial \hat{y}_i}{\partial z_i}+\sum_{j\ne i}\frac{y_j}{1-\hat{y}_{j}}\cdot \frac{\partial \hat{y}_j}{\partial z_i},\\
&\overset{\star}{=}\frac{y_i}{1-\hat{y}_{i}}\cdot \hat{y}_i(1-\hat{y}_i)+\sum_{j\ne i}\frac{y_j}{1-\hat{y}_{j}}\cdot \left(-\hat{y}_i\hat{y}_j\right),\\
&= \left\{\begin{array}{rl}
       \hat{y}_1, & i=1,\\
      -\hat{y}_i\hat{y}_1/\left(1-\hat{y}_1\right), & i\ne 1.
\end{array}\right.
\end{aligned}
\end{equation}
where Equation ($\star$) utilizes the results in Equation~\eqref{equ:softmaxgradient}.

\subsection{Derivatives of KL Div. from Uniform Dist. Loss}\label{a:klu}
The gradient of the KLU loss with respect to $z_i$ is,
\begin{equation}\small
\begin{aligned}
\frac{\partial \mathcal{L}_{\mathrm{KLU}}}{\partial z_i} &=-\sum_{j=1}^{C}\frac{\partial \left(\frac{1}{C}\log\left(C\cdot\hat{y}_{j}\right)\right)}{\partial \hat{y}_j}\cdot \frac{\partial \hat{y}_j}{\partial z_i} \\
&=-\sum_{j=1}^{C}\frac{1}{C\cdot\hat{y}_j}\cdot \frac{\partial \hat{y}_j}{\partial z_i} \\
&\overset{\star}{=}-\frac{1}{C\cdot\hat{y}_i}\cdot\hat{y}_i(1-\hat{y}_i)-\sum_{j\ne i}\frac{1}{C\cdot\hat{y}_j}\cdot \left(-\hat{y}_i\hat{y}_j\right)\\
&= \hat{y}_{i}-\frac{1}{C}.
\end{aligned}
\end{equation}
where Equation ($\star$) utilizes the results in Equation~\eqref{equ:softmaxgradient}.

\begin{figure}[tt]
    \centering
\setlength{\abovecaptionskip}{4pt}
\setlength{\belowcaptionskip}{-5pt}

\includegraphics[width=3.3in, trim=15 10 15 5, clip]{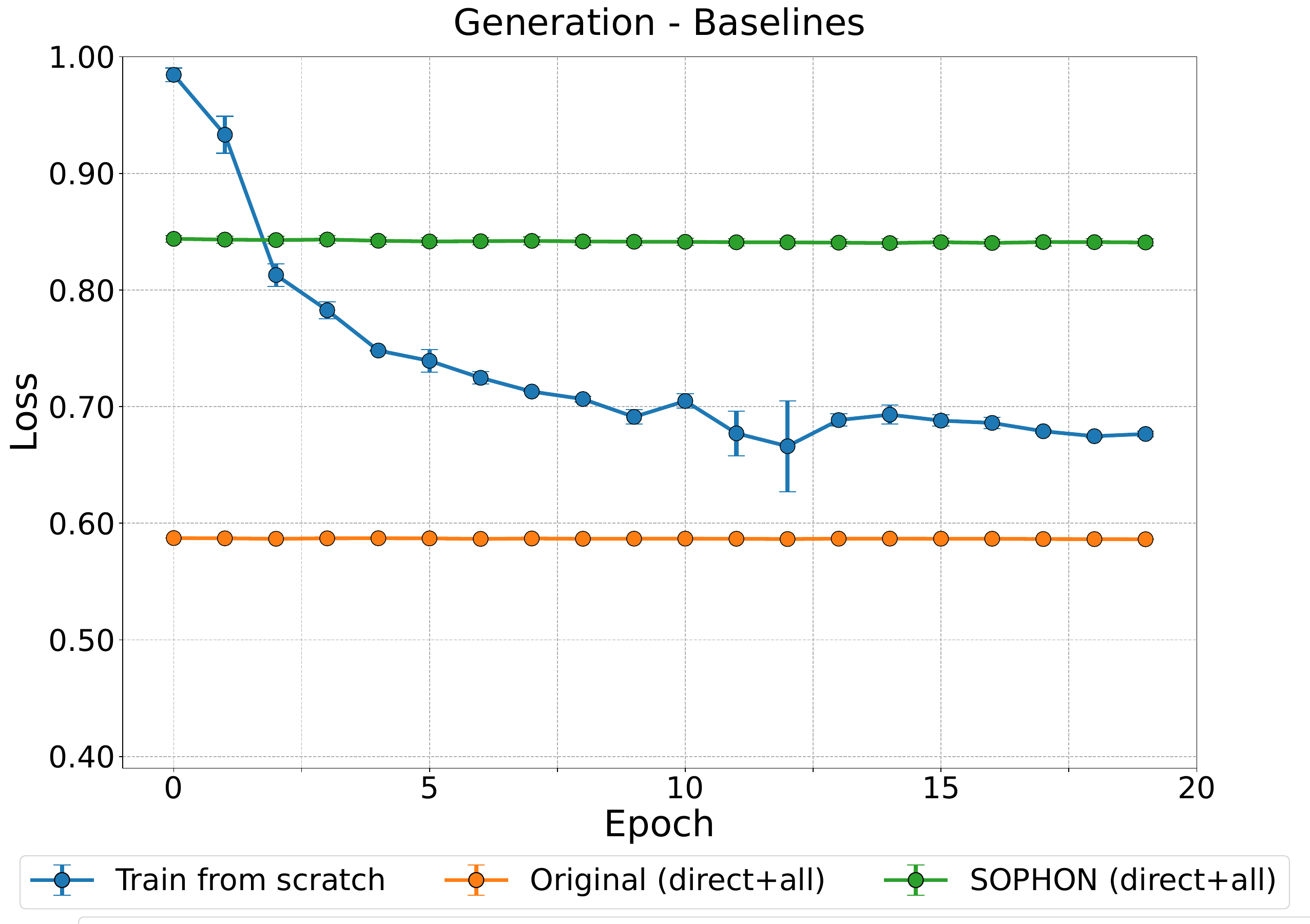}

\caption{Effectiveness of \sys compared with two baselines. The original model can achieve low losses on FFHQ, leading to model misuse. Fine-tuning the \sys model cannot decrease the loss in the restricted domain and yields higher losses than training the model from scratch.}\label{fig:baselines-gen4}
\end{figure}

\begin{table*}\centering
\resizebox{\linewidth}{!}{
\begin{threeparttable}[t]
\setlength{\abovecaptionskip}{0pt}%
\setlength{\belowcaptionskip}{0pt}%
\caption{Effectiveness of \sys against different restricted domains.}

\setlength{\tabcolsep}{5mm}{
\begin{tabular}{@{}lr|c|cccccc@{}}
\toprule
\multirow{2}{*}{Domain}   & \multirow{2}{*}{Method\tnote{\textdagger}} & Original ACC & \multicolumn{6}{c}{ACC in the Restricted Domain}                                 \\
                          &                         & Epoch 0           & Epoch 0   & Epoch 1            & Epoch 5           & Epoch 10          & Epoch 15          & Epoch 20          \\ \midrule
\multirow{3}{*}{CIFAR-10} & Scratch              & -            & -    & \textbf{49.1}$\pm$1.2  & \textbf{59.5}$\pm$0.7 & \textbf{60.6}$\pm$0.5 & \textbf{62.0}$\pm$0.5 & \textbf{62.7}$\pm$0.5 \\
                          & Original              & 99.6         & 9.3 & \textbf{52.4}$\pm$3.8  & \textbf{80.0}$\pm$0.4 & \textbf{83.6}$\pm$0.3 & \textbf{84.6}$\pm$0.2 & \textbf{84.9}$\pm$0.2 \\
                          & \sys                    & 96.2         & 10.0 & \textbf{10.0}$\pm$0.0  & \textbf{10.3}$\pm$0.4 & \textbf{12.8}$\pm$1.5 & \textbf{15.2}$\pm$0.5 & \textbf{14.3}$\pm$1.9 \\ \midrule
\multirow{3}{*}{CINIC}    & Scratch              & -            & -    & \textbf{42.4}$\pm$1.1  & \textbf{50.9}$\pm$0.1 & \textbf{50.2}$\pm$0.7 & \textbf{50.6}$\pm$0.4 & \textbf{51.3}$\pm$0.3 \\
                          & Original              & 99.6         & 9.2 & \textbf{48.9}$\pm$2.8  & \textbf{69.7}$\pm$0.5 & \textbf{72.9}$\pm$0.2 & \textbf{74.2}$\pm$0.2 & \textbf{74.4}$\pm$0.2 \\
                          & \sys                    & 97.6         & 10.0 & \textbf{10.0}$\pm$0.0  & \textbf{9.6}$\pm$0.5  & \textbf{10.1}$\pm$0.3 & \textbf{10.0}$\pm$0.0 & \textbf{8.8}$\pm$1.9  \\ \midrule
\multirow{3}{*}{STL}      & Scratch              & -            & -    & \textbf{23.5}$\pm$4.0  & \textbf{38.9}$\pm$2.5 & \textbf{42.8}$\pm$1.5 & \textbf{45.7}$\pm$0.8 & \textbf{46.3}$\pm$0.3 \\
                          & Original              & 99.6         & 6.9 & \textbf{10.0}$\pm$0.3  & \textbf{19.9}$\pm$2.1 & \textbf{41.6}$\pm$2.6 & \textbf{61.1}$\pm$3.7 & \textbf{71.7}$\pm$5.1 \\
                          & \sys                    & 97.4         & 10.0 & \textbf{10.0}$\pm$0.0  & \textbf{10.0}$\pm$0.0 & \textbf{10.6}$\pm$0.3 & \textbf{11.6}$\pm$0.3 & \textbf{11.0}$\pm$0.2 \\ \midrule
\multirow{3}{*}{MNIST}    & Scratch              & -            & -    & \textbf{87.5}$\pm$16.0 & \textbf{97.9}$\pm$0.6 & \textbf{98.3}$\pm$0.4 & \textbf{98.5}$\pm$0.4 & \textbf{98.4}$\pm$0.3 \\
                          & Original              & 99.6         & 12.3 & \textbf{96.8}$\pm$0.3  & \textbf{98.5}$\pm$0.3 & \textbf{98.9}$\pm$0.1 & \textbf{99.0}$\pm$0.1 & \textbf{99.0}$\pm$0.1 \\
                          & \sys                    & 97.6         & 8.9 & \textbf{11.3}$\pm$0.0  & \textbf{11.3}$\pm$0.0 & \textbf{11.3}$\pm$0.0 & \textbf{11.3}$\pm$0.0 & \textbf{11.8}$\pm$1.0 \\ \midrule
\multirow{3}{*}{SVHN}     & Scratch              & -            & -    & \textbf{44.3}$\pm$4.9  & \textbf{77.7}$\pm$0.4 & \textbf{82.3}$\pm$0.5 & \textbf{84.1}$\pm$0.2 & \textbf{84.8}$\pm$0.2 \\
                          & Original              & 99.6         & 7.4 & \textbf{60.3}$\pm$5.5  & \textbf{90.7}$\pm$0.4 & \textbf{92.6}$\pm$0.3 & \textbf{93.1}$\pm$0.2 & \textbf{93.2}$\pm$0.3 \\
                          & \sys                    & 97.8         & 7.8 & \textbf{19.6}$\pm$0.0  & \textbf{19.6}$\pm$0.0 & \textbf{19.6}$\pm$0.0 & \textbf{19.6}$\pm$0.0 & \textbf{19.6}$\pm$0.0 \\ \bottomrule
\end{tabular}
}

\begin{tablenotes}[flushleft]
\item[\textdagger] Scratch: training from scratch (B1). Original: fine-tuning the original model (B2). \sys: fine-tuning the \sys model.
\end{tablenotes}

\label{tab:domain}
\end{threeparttable}}
\end{table*}
\begin{table*}\centering
\resizebox{\linewidth}{!}{
\begin{threeparttable}[t]
\setlength{\abovecaptionskip}{0pt}%
\setlength{\belowcaptionskip}{0pt}%
\caption{Effectiveness of \sys on different model architectures.}

\setlength{\tabcolsep}{5mm}{
\begin{tabular}{@{}lr|c|cccccc@{}}
\toprule
\multirow{2}{*}{Domain}    & \multirow{2}{*}{Method} & Original ACC & \multicolumn{6}{c}{ACC in the Restricted Domain}                                                                                     \\
                           &                         & Epoch 0      & Epoch 0 & Epoch 1      & Epoch 5      & Epoch 10     & Epoch 15     & Epoch 20      \\ \midrule
\multirow{3}{*}{CAFormer}  & Scratch                 & -            & -    & \textbf{49.1}$\pm$1.2           & \textbf{59.5}$\pm$0.7           & \textbf{60.6}$\pm$0.5            & \textbf{62.0}$\pm$0.5            & \textbf{62.7}$\pm$0.5            \\
                           & Original                & 99.6         & 9.3 & \textbf{52.9}$\pm$3.3           & \textbf{80.0}$\pm$0.5           & \textbf{83.6}$\pm$0.3            & \textbf{84.6}$\pm$0.2            & \textbf{84.9}$\pm$0.2            \\
                           & \sys                    & 96.2         & 10.0 & \textbf{10.9}$\pm$0.6           & \textbf{11.7}$\pm$0.8           & \textbf{11.5}$\pm$1.0            & \textbf{12.0}$\pm$1.1            & \textbf{12.0}$\pm$0.7            \\ \midrule
\multirow{3}{*}{ResNet-50} & Scratch                 & -            & -    & \textbf{37.5}$\pm$4.1           & \textbf{50.4}$\pm$2.6           & \textbf{57.0}$\pm$1.3            & \textbf{60.4}$\pm$0.8            & \textbf{61.5}$\pm$1.1            \\
                           & Original                & 99.2         & 5.2 & \textbf{47.0}$\pm$0.4           & \textbf{66.9}$\pm$0.2           & \textbf{69.9}$\pm$0.2            & \textbf{70.6}$\pm$0.2            & \textbf{70.9}$\pm$0.3            \\
                           & \sys                    & 95.6         & 11.0 & \textbf{11.0}$\pm$0.1           & \textbf{12.6}$\pm$0.3           & \textbf{13.9}$\pm$0.4            & \textbf{15.7}$\pm$0.1            & \textbf{16.7}$\pm$0.1            \\ \midrule
\multirow{3}{*}{ResNet-34} & Scratch                 & -            & -    & \textbf{29.1}$\pm$4.2           & \textbf{57.4}$\pm$3.2           & \textbf{67.5}$\pm$0.9            & \textbf{69.6}$\pm$1.1            & \textbf{70.0}$\pm$0.9            \\
                           & Original                & 99.0         & 5.3 & \textbf{39.9}$\pm$0.1           & \textbf{64.0}$\pm$0.1           & \textbf{70.0}$\pm$0.2            & \textbf{72.4}$\pm$0.3            & \textbf{73.4}$\pm$0.2            \\
                           & \sys                    & 94.8         & 10.0 & \textbf{10.6}$\pm$0.1           & \textbf{11.1}$\pm$0.3           & \textbf{11.1}$\pm$0.4            & \textbf{11.4}$\pm$0.3            & \textbf{11.9}$\pm$0.5            \\ \midrule
\multirow{3}{*}{ResNet-18} & Scratch                 & -            & -    & \textbf{44.8}$\pm$1.7           & \textbf{67.4}$\pm$1.0           & \textbf{71.3}$\pm$0.6            & \textbf{72.5}$\pm$0.5            & \textbf{72.1}$\pm$0.8            \\
                           & Original                & 98.6         & 9.3 & \textbf{41.4}$\pm$0.2           & \textbf{62.4}$\pm$0.2           & \textbf{68.1}$\pm$0.1            & \textbf{70.7}$\pm$0.1            & \textbf{71.5}$\pm$0.3            \\
                           & \sys                    & 95.0         & 8.4 & \textbf{10.7}$\pm$0.1           & \textbf{11.3}$\pm$0.1           & \textbf{11.6}$\pm$0.5            & \textbf{11.6}$\pm$0.2            & \textbf{12.0}$\pm$0.1            \\ \midrule
\multirow{3}{*}{VGG}       & Scratch                 & -            & -    & \textbf{13.7}$\pm$4.6           & \textbf{58.0}$\pm$1.0           & \textbf{75.0}$\pm$1.1            & \textbf{78.0}$\pm$0.6            & \textbf{78.8}$\pm$0.3            \\
                           & Original                & 95.4         & 2.8 & \textbf{33.9}$\pm$0.2           & \textbf{72.4}$\pm$0.2           & \textbf{79.9}$\pm$0.3            & \textbf{82.9}$\pm$0.2            & \textbf{84.6}$\pm$0.2            \\
                           & \sys                    & 91.2         & 10.5 & \textbf{10.4}$\pm$0.1           & \textbf{10.9}$\pm$0.3           & \textbf{10.8}$\pm$0.8            & \textbf{17.4}$\pm$4.2            & \textbf{18.0}3$\pm$0.7           \\ \bottomrule
\end{tabular}
}

\begin{tablenotes}[flushleft]
\item[\textdagger] Scratch: training from scratch (B1). Original: fine-tuning the original model (B2). \sys: fine-tuning the \sys model.
\end{tablenotes}

\label{tab:model}
\end{threeparttable}}
\end{table*}
\begin{table*}\centering
\resizebox{\linewidth}{!}{
\begin{threeparttable}[t]
\setlength{\abovecaptionskip}{0pt}%
\setlength{\belowcaptionskip}{0pt}%
\caption{Effectiveness of \sys against different optimizers.}

\setlength{\tabcolsep}{5mm}{
\begin{tabular}{@{}lr|c|cccccc@{}}
\toprule
\multirow{2}{*}{Optimizer} & \multirow{2}{*}{Method} & Original ACC & \multicolumn{6}{c}{ACC in the Restricted Domain}                                     \\
                           &                         & Epoch 0      & Epoch 0 & Epoch 1      & Epoch 5      & Epoch 10     & Epoch 15      & Epoch 20      \\ \midrule
\multirow{3}{*}{Momentum}  & Scratch                 & -            & -       & \textbf{49.3}$\pm$1.0 & \textbf{59.4}$\pm$0.7 & \textbf{60.3}$\pm$0.2 & \textbf{61.7}$\pm$0.6  & \textbf{62.2}$\pm$0.7  \\
                           & Original                & 99.6         & 9.3     & \textbf{51.5}$\pm$4.5 & \textbf{79.7}$\pm$1.1 & \textbf{83.4}$\pm$0.6 & \textbf{84.8}$\pm$0.2  & \textbf{85.0}$\pm$0.4  \\
                           & \sys                    & 96.2         & 10.0    & \textbf{10.0}$\pm$0.0 & \textbf{10.0}$\pm$0.0 & \textbf{10.2}$\pm$0.7 & \textbf{10.9}$\pm$0.7  & \textbf{10.6}$\pm$0.8  \\ \midrule
\multirow{3}{*}{Nesterov}  & Scratch                 & -            & -       & \textbf{51.1}$\pm$0.8 & \textbf{59.8}$\pm$0.6 & \textbf{61.1}$\pm$0.6 & \textbf{61.9}$\pm$0.2  & \textbf{62.6}$\pm$0.7  \\
                           & Original                & 99.6         & 9.3     & \textbf{50.1}$\pm$3.8 & \textbf{79.8}$\pm$0.3 & \textbf{83.4}$\pm$0.2 & \textbf{84.7}$\pm$0.2  & \textbf{84.9}$\pm$0.2  \\
                           & \sys                    & 96.2         & 10.0    & \textbf{10.0}$\pm$0.0 & \textbf{10.0}$\pm$0.0 & \textbf{10.1}$\pm$0.7 & \textbf{10.9}$\pm$0.7  & \textbf{10.4}$\pm$0.9  \\ \midrule
\multirow{3}{*}{Adagrad}   & Scratch                 & -            & -       & \textbf{47.8}$\pm$1.2 & \textbf{62.6}$\pm$0.4 & \textbf{64.3}$\pm$0.3 & \textbf{64.3}$\pm$0.3  & \textbf{64.2}$\pm$0.4  \\
                           & Original                & 99.6         & 9.3     & \textbf{80.1}$\pm$1.1 & \textbf{84.8}$\pm$0.5 & \textbf{84.9}$\pm$0.2 & \textbf{84.7}$\pm$0.4  & \textbf{84.6}$\pm$0.4  \\
                           & \sys                    & 96.2         & 10.0    & \textbf{10.0}$\pm$0.0 & \textbf{10.0}$\pm$0.0 & \textbf{9.9}$\pm$0.1  & \textbf{12.0}$\pm$2.4  & \textbf{22.2}$\pm$12.7 \\ \midrule
\multirow{3}{*}{Adadelta}  & Scratch                 & -            & -       & \textbf{49.5}$\pm$0.8 & \textbf{59.7}$\pm$0.5 & \textbf{60.8}$\pm$0.7 & \textbf{63.6}$\pm$1.0  & \textbf{64.4}$\pm$0.4  \\
                           & Original                & 99.6         & 9.3     & \textbf{80.1}$\pm$0.6 & \textbf{86.6}$\pm$0.4 & \textbf{87.1}$\pm$0.2 & \textbf{87.3}$\pm$0.3  & \textbf{87.3}$\pm$0.3  \\
                           & \sys                    & 96.2         & 10.0    & \textbf{9.8}$\pm$0.5  & \textbf{12.3}$\pm$2.5 & \textbf{17.6}$\pm$1.0 & \textbf{31.4}$\pm$4.1  & \textbf{56.4}$\pm$2.1  \\ \midrule
\multirow{3}{*}{Adam}      & Scratch                 & -            & -       & \textbf{53.7}$\pm$0.2 & \textbf{62.7}$\pm$0.4 & \textbf{63.4}$\pm$0.6 & \textbf{63.8}$\pm$0.2  & \textbf{63.5}$\pm$0.5  \\
                           & Original                & 99.6         & 9.3     & \textbf{87.4}$\pm$0.9 & \textbf{89.8}$\pm$0.3 & \textbf{90.1}$\pm$0.3 & \textbf{90.0}$\pm$0.4  & \textbf{89.9}$\pm$0.3  \\
                           & \sys                    & 96.2         & 10.0    & \textbf{10.0}$\pm$0.0 & \textbf{9.8}$\pm$0.4  & \textbf{10.0}$\pm$0.0 & \textbf{23.2}$\pm$10.8 & \textbf{50.2}$\pm$4.0  \\ \bottomrule
\end{tabular}
}

\begin{tablenotes}[flushleft]
\item[\textdagger] Scratch: training from scratch (B1). Original: fine-tuning the original model (B2). \sys: fine-tuning the \sys model.
\end{tablenotes}

\label{tab:optimizer}
\end{threeparttable}}
\end{table*}
\begin{table*}\centering
\resizebox{\linewidth}{!}{
\begin{threeparttable}[t]
\setlength{\abovecaptionskip}{0pt}%
\setlength{\belowcaptionskip}{0pt}%
\caption{Effectiveness of \sys against different learning rates.}

\setlength{\tabcolsep}{5mm}{
\begin{tabular}{@{}lr|c|cccccc@{}}
\toprule
\multirow{2}{*}{\begin{tabular}[c]{@{}l@{}}Learning\\ Rate\end{tabular}} & \multirow{2}{*}{Method} & Original ACC & \multicolumn{6}{c}{ACC in the Restricted Domain}                                     \\
                                                                         &                         & Epoch 0      & Epoch 0 & Epoch 1      & Epoch 5      & Epoch 10     & Epoch 15      & Epoch 20      \\ \midrule
\multirow{3}{*}{$5\times 10^{-6}$}                                       & Scratch                 & -            & -       & \textbf{20.3}$\pm$1.2 & \textbf{31.7}$\pm$0.2 & \textbf{35.9}$\pm$0.0 & \textbf{38.0}$\pm$0.1  & \textbf{39.7}$\pm$0.1  \\
                                                                         & Original                & 99.6         & 9.3     & \textbf{11.1}$\pm$0.3 & \textbf{20.9}$\pm$0.4 & \textbf{37.8}$\pm$0.4 & \textbf{55.3}$\pm$0.2  & \textbf{63.0}$\pm$0.1  \\
                                                                         & \sys                    & 96.2         & 10.0    & \textbf{1.4}$\pm$0.0  & \textbf{1.3}$\pm$0.0  & \textbf{1.2}$\pm$0.0  & \textbf{1.1}$\pm$0.0   & \textbf{1.0}$\pm$0.0   \\ \midrule
\multirow{3}{*}{$1\times 10^{-5}$}                                       & Scratch                 & -            & -       & \textbf{25.0}$\pm$0.9 & \textbf{35.8}$\pm$0.1 & \textbf{39.6}$\pm$0.1 & \textbf{41.9}$\pm$0.3  & \textbf{43.5}$\pm$0.2  \\
                                                                         & Original                & 99.6         & 9.3     & \textbf{13.1}$\pm$0.3 & \textbf{37.1}$\pm$0.6 & \textbf{62.6}$\pm$0.4 & \textbf{70.3}$\pm$0.4  & \textbf{74.0}$\pm$0.3  \\
                                                                         & \sys                    & 96.2         & 10.0    & \textbf{1.4}$\pm$0.0  & \textbf{1.2}$\pm$0.0  & \textbf{1.0}$\pm$0.0  & \textbf{0.9}$\pm$0.0   & \textbf{0.9}$\pm$0.0   \\ \midrule
\multirow{3}{*}{$1\times 10^{-4}$}                                       & Scratch                 & -            & -       & \textbf{39.2}$\pm$0.4 & \textbf{48.2}$\pm$0.2 & \textbf{52.3}$\pm$0.2 & \textbf{54.0}$\pm$0.3  & \textbf{54.9}$\pm$0.3  \\
                                                                         & Original                & 99.6         & 9.3     & \textbf{53.0}$\pm$1.0 & \textbf{79.8}$\pm$0.2 & \textbf{83.7}$\pm$0.1 & \textbf{84.6}$\pm$0.1  & \textbf{84.8}$\pm$0.2  \\
                                                                         & \sys                    & 96.2         & 10.0    & \textbf{1.0}$\pm$0.0  & \textbf{1.4}$\pm$0.0  & \textbf{8.5}$\pm$0.4  & \textbf{9.8}$\pm$0.1   & \textbf{8.9}$\pm$1.1   \\ \midrule
\multirow{3}{*}{$1\times 10^{-3}$}                                       & Scratch                 & -            & -       & \textbf{48.8}$\pm$0.3 & \textbf{56.5}$\pm$0.3 & \textbf{58.4}$\pm$0.1 & \textbf{58.8}$\pm$0.0  & \textbf{58.7}$\pm$0.0  \\
                                                                         & Original                & 99.6         & 9.3     & \textbf{74.1}$\pm$0.6 & \textbf{86.0}$\pm$0.7 & \textbf{86.8}$\pm$0.7 & \textbf{87.2}$\pm$0.5  & \textbf{87.2}$\pm$1.0  \\
                                                                         & \sys                    & 96.2         & 10.0    & \textbf{8.0}$\pm$2.5  & \textbf{10.0}$\pm$0.0 & \textbf{10.0}$\pm$0.0 & \textbf{10.0}$\pm$0.0  & \textbf{10.0}$\pm$0.0  \\ \midrule
\multirow{3}{*}{$1\times 10^{-2}$}                                       & Scratch                 & -            & -       & \textbf{50.2}$\pm$1.2 & \textbf{59.7}$\pm$0.6 & \textbf{60.5}$\pm$0.4 & \textbf{62.1}$\pm$0.6  & \textbf{62.8}$\pm$0.4  \\
                                                                         & Original                & 99.6         & 9.3     & \textbf{10.0}$\pm$0.0 & \textbf{10.0}$\pm$0.0 & \textbf{10.0}$\pm$0.0 & \textbf{23.5}$\pm$13.5 & \textbf{41.9}$\pm$31.9 \\
                                                                         & \sys                    & 96.2         & 10.0    & \textbf{10.0}$\pm$0.0 & \textbf{10.0}$\pm$0.0 & \textbf{10.0}$\pm$0.0 & \textbf{10.0}$\pm$0.0  & \textbf{39.2}$\pm$34.8 \\ \bottomrule
\end{tabular}
}

\begin{tablenotes}[flushleft]
\item[\textdagger] Scratch: training from scratch (B1). Original: fine-tuning the original model (B2). \sys: fine-tuning the \sys model.
\end{tablenotes}

\label{tab:lr}
\end{threeparttable}}
\end{table*}
\begin{table*}\centering
\resizebox{\linewidth}{!}{
\begin{threeparttable}[t]
\setlength{\abovecaptionskip}{0pt}%
\setlength{\belowcaptionskip}{0pt}%
\caption{Effectiveness of \sys against different batch sizes.}

\setlength{\tabcolsep}{5mm}{
\begin{tabular}{@{}lr|c|cccccc@{}}
\toprule
\multirow{2}{*}{\begin{tabular}[c]{@{}l@{}}Batch\\ Size\end{tabular}} & \multirow{2}{*}{Method} & Original ACC & \multicolumn{6}{c}{ACC in the Restricted Domain}                                                                                                                 \\
                                                                      &                         & Epoch 0      & Epoch 0 & \multicolumn{1}{c}{Epoch 1} & \multicolumn{1}{c}{Epoch 5} & \multicolumn{1}{c}{Epoch 10} & \multicolumn{1}{c}{Epoch 15} & \multicolumn{1}{c}{Epoch 20} \\ \midrule
\multirow{3}{*}{50}                                                   & Scratch                 & -            & -       & \textbf{46.2}$\pm$0.3                & \textbf{54.6}$\pm$0.4                & \textbf{55.9}$\pm$0.2                 & \textbf{56.4}$\pm$0.2                 & \textbf{57.0}$\pm$0.3                 \\
                                                                      & Original                & 99.6         & 9.3     & \textbf{72.9}$\pm$2.9                & \textbf{84.7}$\pm$0.2                & \textbf{86.1}$\pm$0.5                 & \textbf{86.5}$\pm$0.1                 & \textbf{86.6}$\pm$0.5                 \\
                                                                      & \sys                    & 96.2         & 10.0    & \textbf{1.1}$\pm$0.1                 & \textbf{8.2}$\pm$1.2                 & \textbf{10.0}$\pm$0.0                 & \textbf{9.2}$\pm$1.4                  & \textbf{9.9}$\pm$0.0                  \\ \midrule
\multirow{3}{*}{100}                                                  & Scratch                 & -            & -       & \textbf{42.8}$\pm$0.4                & \textbf{51.8}$\pm$0.3                & \textbf{54.9}$\pm$0.1                 & \textbf{55.2}$\pm$0.1                 & \textbf{55.3}$\pm$0.4                 \\
                                                                      & Original                & 99.6         & 9.3     & \textbf{63.0}$\pm$3.9                & \textbf{83.1}$\pm$0.3                & \textbf{85.4}$\pm$0.1                 & \textbf{85.7}$\pm$0.2                 & \textbf{85.7}$\pm$0.2                 \\
                                                                      & \sys                    & 96.2         & 10.0    & \textbf{0.9}$\pm$0.0                 & \textbf{8.2}$\pm$1.0                 & \textbf{9.4}$\pm$0.8                  & \textbf{7.7}$\pm$2.1                  & \textbf{9.2}$\pm$1.2                  \\ \midrule
\multirow{3}{*}{150}                                                  & Scratch                 & -            & -       & \textbf{40.5}$\pm$0.3                & \textbf{49.7}$\pm$0.1                & \textbf{53.6}$\pm$0.2                 & \textbf{54.9}$\pm$0.1                 & \textbf{55.2}$\pm$0.5                 \\
                                                                      & Original                & 99.6         & 9.3     & \textbf{59.4}$\pm$1.9                & \textbf{81.3}$\pm$0.1                & \textbf{84.6}$\pm$0.2                 & \textbf{85.2}$\pm$0.2                 & \textbf{85.2}$\pm$0.2                 \\
                                                                      & \sys                    & 96.2         & 10.0    & \textbf{0.9}$\pm$0.0                 & \textbf{1.6}$\pm$0.1                 & \textbf{9.8}$\pm$0.0                  & \textbf{8.4}$\pm$1.1                  & \textbf{8.2}$\pm$1.6                  \\ \midrule
\multirow{3}{*}{200}                                                  & Scratch                 & -            & -       & \textbf{39.2}$\pm$0.4                & \textbf{48.2}$\pm$0.2                & \textbf{52.3}$\pm$0.2                 & \textbf{54.0}$\pm$0.3                 & \textbf{54.9}$\pm$0.3                 \\
                                                                      & Original                & 99.6         & 9.3     & \textbf{53.0}$\pm$1.0                & \textbf{79.8}$\pm$0.2                & \textbf{83.7}$\pm$0.1                 & \textbf{84.6}$\pm$0.1                 & \textbf{84.8}$\pm$0.2                 \\
                                                                      & \sys                    & 96.2         & 10.0    & \textbf{1.0}$\pm$0.0                 & \textbf{1.4}$\pm$0.0                 & \textbf{8.3}$\pm$0.4                  & \textbf{9.8}$\pm$0.1                  & \textbf{8.1}$\pm$1.3                  \\ \midrule
\multirow{3}{*}{250}                                                  & Scratch                 & -            & -       & \textbf{38.0}$\pm$0.3                & \textbf{46.9}$\pm$0.2                & \textbf{51.1}$\pm$0.1                 & \textbf{53.1}$\pm$0.3                 & \textbf{54.0}$\pm$0.2                 \\
                                                                      & Original                & 99.6         & 9.3     & \textbf{41.2}$\pm$2.7                & \textbf{78.4}$\pm$0.3                & \textbf{82.9}$\pm$0.3                 & \textbf{84.1}$\pm$0.1                 & \textbf{84.8}$\pm$0.1                 \\
                                                                      & \sys                    & 96.2         & 10.0    & \textbf{1.0}$\pm$0.0                 & \textbf{1.2}$\pm$0.0                 & \textbf{3.6}$\pm$0.2                  & \textbf{9.8}$\pm$0.0                  & \textbf{9.9}$\pm$0.0                  \\ \bottomrule
\end{tabular}
}

\begin{tablenotes}[flushleft]
\item[\textdagger] Scratch: training from scratch (B1). Original: fine-tuning the original model (B2). \sys: fine-tuning the \sys model.
\end{tablenotes}

\label{tab:batchsize}
\end{threeparttable}}
\end{table*}


\clearpage
\balance
\section{Meta-Review}

The following meta-review was prepared by the program committee for the 2024 IEEE Symposium on Security and Privacy (S\&P) as part of the review process as detailed in the call for papers.

\subsection{Summary}
This paper proposes a framework (named SOPHON) for model pre-training. The goal is to prevent pre-trained models from being fine-tuned to “unintended” use cases while preserving the performance of the original use case. The intuition is to trap the pre-trained model within a hard-to-escape local optimum of the restricted domains. At the core, SOPHON combines two loss terms designed to i) preserve the model's general capabilities and ii) reduce its capabilities for particular domains via fine-tuning simulation, which was inspired by model-agnostic meta-learning paradigms. Experimental evaluation showed SOPHON’s success against various models and datasets in vision domains.

\subsection{Scientific Contributions}
\noindent Provides a Valuable Step Forward in an Established Field.

\subsection{Reasons for Acceptance}
\begin{enumerate}[leftmargin=20pt]
\item Investigates an important problem -- ``non-fine-tunability'', a generalized case of ``non-transferability''.
\item The system supports both classification and generative models.
\item The proposed approach nicely utilizes the insights from relevant literature to realize the goal.
\end{enumerate}

\subsection{Noteworthy Concerns} 
\begin{enumerate}[leftmargin=20pt]
\item In all settings, the worst penalty the attacker can pay is training from scratch. So, an increase in accuracy compared to training from scratch is a partial success for the attacker.
\item The paper considers a partial set of training regimes/domain-adaptation techniques. How the results would generalize for the untested/unseen domains is unclear.
\end{enumerate}



\end{document}